\documentclass{article}

\usepackage{microtype}
\usepackage{graphicx}
\usepackage{subfigure}
\usepackage{booktabs} % for professional tables

\usepackage{hyperref}

\usepackage[accepted]{icml2025}

\usepackage{amsmath}
\usepackage{amssymb}
\usepackage{mathtools}
\usepackage{amsthm}
\usepackage{mdframed}
\usepackage{multicol,multirow}
\usepackage{colortbl}
\usepackage{caption}
\usepackage[most]{tcolorbox}

\usepackage[capitalize,noabbrev]{cleveref}

\newmdenv[
    backgroundcolor=blue!20, % Light gray background
    linecolor=blue!20,       % Border color (hidden)
    innerleftmargin=10pt,
    innerrightmargin=10pt,
    innertopmargin=10pt,
    innerbottommargin=10pt,
    skipabove=10pt,
    skipbelow=10pt
]{theoremframe}

\newmdenv[
    backgroundcolor=gray!20, % Light gray background
    linecolor=gray!20,       % Border color (hidden)
    innerleftmargin=10pt,
    innerrightmargin=10pt,
    innertopmargin=10pt,
    innerbottommargin=10pt,
    skipabove=10pt,
    skipbelow=10pt
]{assumptionframe}

\theoremstyle{plain}
\newtheorem{theorem}{Theorem}[section]

\theoremstyle{definition}

\newtheorem{assumption}[theorem]{Assumption}
\theoremstyle{remark}

\newenvironment{framedtheorem}
    {\begin{theoremframe}\begin{theorem}}
    {\end{theorem}\end{theoremframe}}
\newenvironment{framedassumption}
    {\begin{assumptionframe}\begin{assumption}}
    {\end{assumption}\end{assumptionframe}}

\usepackage[textsize=tiny]{todonotes}

\icmltitlerunning{Ultra-Resolution Adaptation with Ease}

\begin{document}

% \twocolumn[

\twocolumn[{%
    \renewcommand\twocolumn[1][]{#1}%
    \icmltitle{Ultra-Resolution Adaptation with Ease}
    \icmlsetsymbol{equal}{*}

\begin{icmlauthorlist}
\icmlauthor{Ruonan Yu}{equal,yyy}
\icmlauthor{Songhua Liu}{equal,yyy}
\icmlauthor{Zhenxiong Tan}{yyy}
\icmlauthor{Xinchao Wang}{yyy}
\end{icmlauthorlist}
    \begin{center}
        \includegraphics[width=\textwidth]{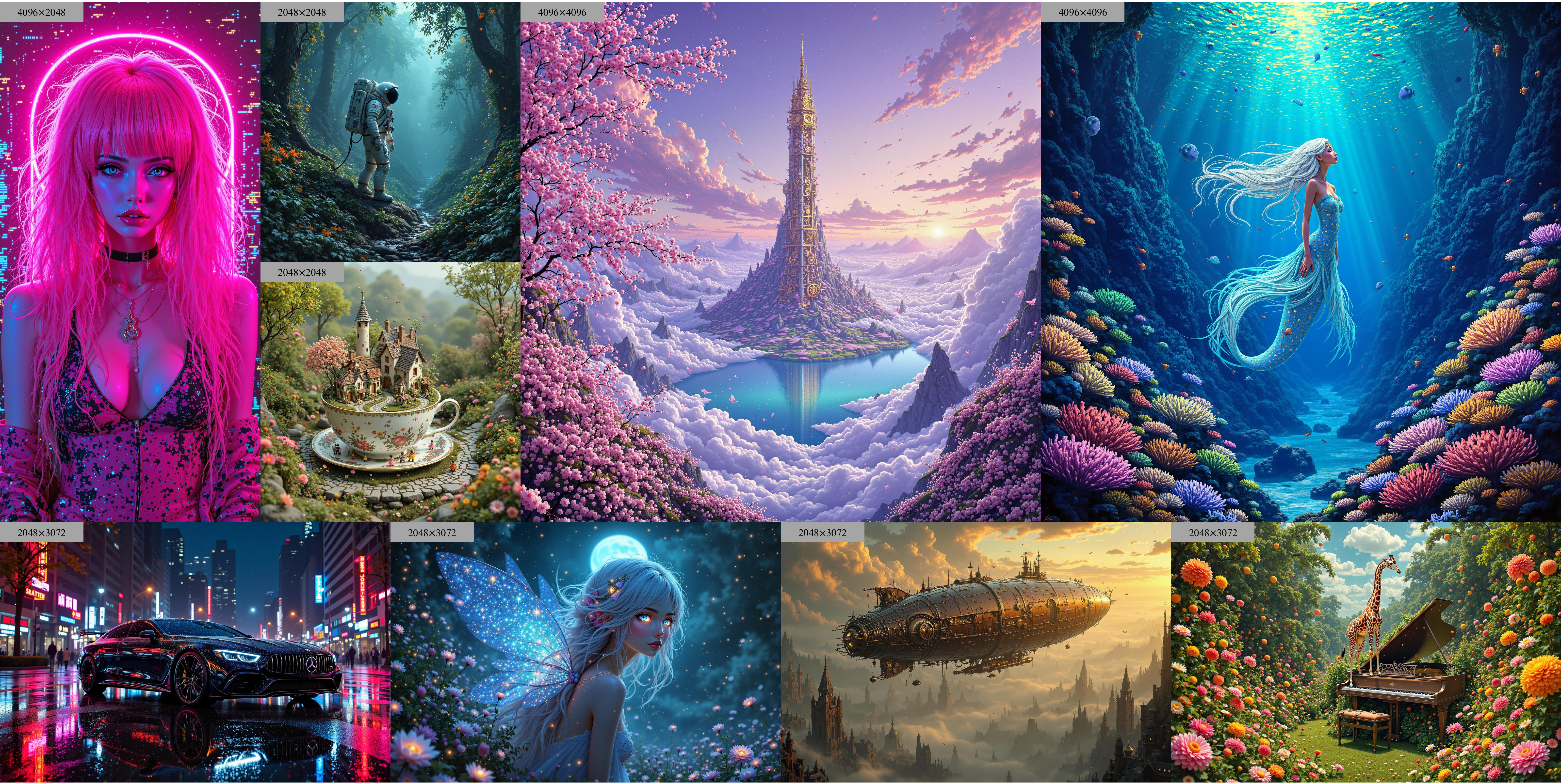}
        \captionof{figure}{High-resolution results by our method.}
        \label{fig:main}
    \end{center}
    \vspace{-0.4cm}
}



\icmlaffiliation{yyy}{National University of Singapore}

\icmlcorrespondingauthor{Xinchao Wang}{xinchao@nus.edu.sg}

\icmlkeywords{Ultra-Resolution Generation, Efficient Tuning, Synthetic Data, Diffusion Models}

\vskip 0.3in
]



\printAffiliationsAndNotice{\icmlEqualContribution} % otherwise use the standard text.

\begin{abstract}
Text-to-image diffusion models have achieved remarkable progress in recent years. 
However, training models for high-resolution image generation remains challenging, particularly when training data and computational resources are limited. 
In this paper, we explore this practical problem from two key perspectives: data and parameter efficiency, and propose a set of key guidelines for ultra-resolution adaptation termed \emph{URAE}. 
For data efficiency, we theoretically and empirically demonstrate that synthetic data generated by some teacher models can significantly promote training convergence. 
For parameter efficiency, we find that tuning minor components of the weight matrices outperforms widely-used low-rank adapters when synthetic data are unavailable, offering substantial performance gains while maintaining efficiency. 
Additionally, for models leveraging guidance distillation, such as FLUX, we show that disabling classifier-free guidance, \textit{i.e.}, setting the guidance scale to 1 during adaptation, is crucial for satisfactory performance. 
Extensive experiments validate that URAE achieves comparable 2K-generation performance to state-of-the-art closed-source models like FLUX1.1 [Pro] Ultra with only 3K samples and 2K iterations, while setting new benchmarks for 4K-resolution generation. 
Codes are available \href{https://github.com/Huage001/URAE}{here}. 
\end{abstract}

\section{Introduction}

Recent years have witnessed remarkable progress in text-to-image generation with diffusion models~\cite{nichol2021improved,dhariwal2021diffusion,rombach2022high,ho2020denoising}. 
From UNet-based architectures~\cite{ronneberger2015u,rombach2022high} to latest state-of-the-art Diffusion Transformers (DiTs)~\cite{peebles2023scalable,bao2023all,chen2023pixart,esser2024scaling,li2024hunyuan,gao2024lumina,chen2024pixartsigma}, these models leverage powerful backbones and multistep denoising schemes to generate high-quality and diverse images from textual prompts effectively, solidifying their leading position in this field~\cite{croitoru2023diffusion,yang2023diffusion}. 

Nevertheless, extending current diffusion models to ultra-resolution generation, such as 4K, remains a significant challenge. 
The process typically demands massive amounts of high-quality data and substantial computational resources, making training at such resolutions daunting and accessible only to industry-scale efforts. 
Although recent attempts have been made to train 4K-resolution text-to-image models~\cite{chen2024pixartsigma,xie2024sana}, they rely on internal datasets containing millions of high-resolution images to fine-tune base low-resolution models. 
In practice, collecting such large-scale datasets for training is highly cumbersome if not infeasible at all. 
Meanwhile, tuning the entire diffusion backbone introduces an intensive GPU memory footprint, especially for state-of-the-art models like FLUX~\cite{flux2023} and Stable Diffusion 3.5~\cite{esser2024scaling}. 

Focusing on these drawbacks, we are curious about one practical question: \emph{Can this ultra-resolution adaptation process be made easier?} 
In this paper, we answer the question positively by proposing \emph{URAE}, a set of key guidelines, under which ultra-resolution adaptation is achievable with merely thousands of training samples and iterations. 

Specifically, we initiate our exploration from two key aspects: data and parameter efficiency. 
On the one hand, we provide theoretical and empirical evidence that synthetic data produced by some teacher models can largely enhance training convergence. 
However, despite recent advancements in text-to-image generation, state-of-the-art models still face significant challenges in acquiring high-quality synthetic training data for ultra-resolution adaptation, such as 4K. 
We thus, on the other hand, investigate such scenarios where synthetic data are unavailable and identify that tuning minor components of the pre-trained weight matrices is more effective than commonly used parameter-efficient adaptation strategies like LoRA~\cite{hu2022lora}. 

Furthermore, we delve into the principles of fine-tuning guidance-distilled models like FLUX and discover that disabling classifier-free guidance—by setting the guidance scale to 1—is essential, regardless of the availability of synthetic data. 
Backed up by the above guidelines, we conduct extensive experiments to demonstrate that URAE achieves performance comparable to state-of-the-art closed-source models like FLUX1.1 [Pro] Ultra with merely 3K training samples and 2K adaptation iterations. 
Meanwhile, it surpasses previous models in 4K generation performance and remains highly compatible with existing training-free high-resolution generation pipelines~\cite{du2024max,meng2021sdedit}, enabling further performance improvements. 
In summary, the contributions of this paper are:
\begin{itemize}
    \item We are the first to delve into the problem of ultra-resolution adaption to the best of our knowledge;
    \item We propose URAE, a set of key guidelines focusing on data efficiency, parameter efficiency, and classifier-free guidance, to facilitate the adaptation of existing text-to-image models to higher resolutions; 
    \item We validate that URAE achieves comparable performance in 2K generation, superior capabilities in 4K generation, and strong compatibility with existing training-free high-resolution generation pipelines. 
    % Codes will be released to facilitate future research. 
\end{itemize}

\section{Methodology}

In this section, we delve into the motivations and technical details of URAE, our proposed strategy for ultra-resolution adaptation. 
We begin with some preliminary concepts, followed by three key components including training with synthetic data, parameter-efficient fine-tuning strategies, and classifier-free guidance. 

\subsection{Preliminary}

State-of-the-art text-to-image diffusion models commonly adopt the flow matching training scheme~\cite{esser2024scaling, lipman2022flow}. 
Specifically, in each iteration, a batch of images $x$ and their corresponding textual descriptions $y$ is sampled. 
These images are then encoded into a latent map $z_0$ using a pre-trained VAE encoder, and a noise map $\epsilon$ is drawn from a Gaussian distribution. 
Let $z_t$ denote the noisy version of $z_0$ after applying $\epsilon$ at the $t$-th diffusion timestep. 
The flow matching loss is then formulated as: 
\begin{equation}
    \mathcal{L}_{fm}(z_0,y,t,\epsilon)=\Vert(\epsilon-z_0)-\epsilon_\theta(z_t,t,y)\Vert_2^2,\label{eq:1}
\end{equation}
where $\epsilon_\theta(\cdot)$ is the denoising backbone with parameters $\theta$. 

The inference process begins with a text prompt $y$ and a random Gaussian noise $\epsilon$, also denoted $z_T$, which is iteratively denoised using the trained backbone. 
After $T$ steps, the resulting $z_0$ represents a clean sample in the latent space and is then decoded to a generated image $x$ using a pre-trained VAE decoder. 
Such training and inference paradigms are also employed in our URAE framework. 

\subsection{Synthetic Data or Real Data?}

Previous works train 4K-generation models using millions of high-quality training images~\cite{chen2024pixartsigma,xie2024sana}, leading to significant challenges in collecting, transmitting, storing, and processing such large volumes of data. 
To alleviate these inconveniences, we target a data-efficient approach for ultra-resolution adaptation. 

Building on recent advances in the distillation of diffusion models~\cite{xingyi2023diffusion, kim2023bk, liu2024linfusion, liu2024clear}, we recognize that incorporating a teacher model for reference and a loss term for knowledge distillation~\cite{hinton2015distilling} can enhance training: 
\begin{equation}
    \mathcal{L}_{distill}(z_0,y,t,\epsilon)=\Vert\epsilon_{\theta}(z_t,t,y)-\epsilon_{\theta_{ref}}(z_t,t,y)\Vert_2^2,
\end{equation} 
where $\theta_{ref}$ represents the parameters of the teacher model. 
However, this approach relies on access to the diffusion backbone of the teacher model to compute the step-wise distillation loss, which is impractical for closed-weight models such as FLUX1.1 [Pro] Ultra. 
We therefore experiment with an alternative approach that optimizes the vanilla flow matching loss defined in Eq.~\ref{eq:1} using data synthesized by the teacher model. 
We expect this to yield similar training benefits, as validated by the following theoretical analysis. 

Before presenting our main results, we first set up some necessary assumptions. 
From the model perspective: 
\begin{framedassumption}
Let $u$ denote the input data pair $(\epsilon,y)$. 
The process from $u$ to the output $x$ can be characterized by a neural network $f(u;W)$ with infinite width, where $W$ denotes the network's parameters. \label{ass:1}
\end{framedassumption}
It has been demonstrated that neural networks with a single hidden layer and sufficient width can approximate any complex functions~\cite{cybenko1989approximation,hornik1991approximation}, and that infinite-width neural networks trained with gradient descent are equivalent to training linear models in the space of Neural Tangent Kernel (NTK)~\cite{jacot2018neural}. 
From the data perspective: 
\begin{framedassumption}
The dataset contains both real and synthetic data, with the synthetic portion denoted as $p\in[0,1]$. 
The total number of data is $N$. 
Given an input $u$, the corresponding target $x$ in the real data distribution is characterized by $x_{real} = f(u; W^*) + \xi$, where $\xi \sim \mathcal{N}(0, \sigma^2)$ is a scaler from a noise distribution and $W^*$ denotes the optimal parameters, serving as an unknown oracle. 
For synthetic data, the target is given by $x_{syn}=f(u;W_{ref})$, where $W_{ref}$ denotes the parameters of a pre-trained teacher model for reference. 
Without loss of generality, we restrict our discussion to the one-dimensional target case. \label{ass:2}
\end{framedassumption}
And from the training perspective:
\begin{framedassumption}
The model's parameters are initialized as $W_0$. 
The training is conducted through SGD using the square error as the loss function: $\mathcal{L}=\frac{1}{2N}\Vert f(u;W)-x\Vert_2^2$. 
The learning rate is $\eta$. \label{ass:3}
\end{framedassumption}
Our main results are summarized in Theorem~\ref{thm:1}:
\begin{framedtheorem}
Under the setting defined in Assumptions~\ref{ass:1}, \ref{ass:2}, and \ref{ass:3}, the error between $W_T$, the parameters after $T$ training iterations, and the optimal $W^*$ is bounded by:
\begin{equation}
    \begin{aligned}
    &\mathbb{E}[\Vert W_T - W^*\Vert_2^2]\leq\mathbb{E}[\Vert(I - \eta M)^T \Delta_0\Vert_2^2]+
    \\&\eta^2 (p(1-p)\mathbb{E}[\delta^2]+(1-p)\sigma^2) \sum_{i=1}^{N} \frac{(1 - (1 - \eta \lambda_i)^T)^2}{\lambda_i}\\&+p^2\Vert W_{ref}-W^*\Vert_2^2.
    \end{aligned}
\end{equation}\label{eq:3}
where $\Delta_0=W_0-(pW_{ref}+(1-p)W^*)$, $M$ is defined as $\nabla_{W}f(U;W_0)^\top\nabla_{W}f(U;W_0)$, $\delta=f(u;W_{ref})-f(u;W^*)$, and $\lambda_i$ is the $i$-th eigenvalue of $M$. 
\label{thm:1}
\end{framedtheorem}

\begin{figure}[t]
  \centering
   \includegraphics[width=\linewidth]{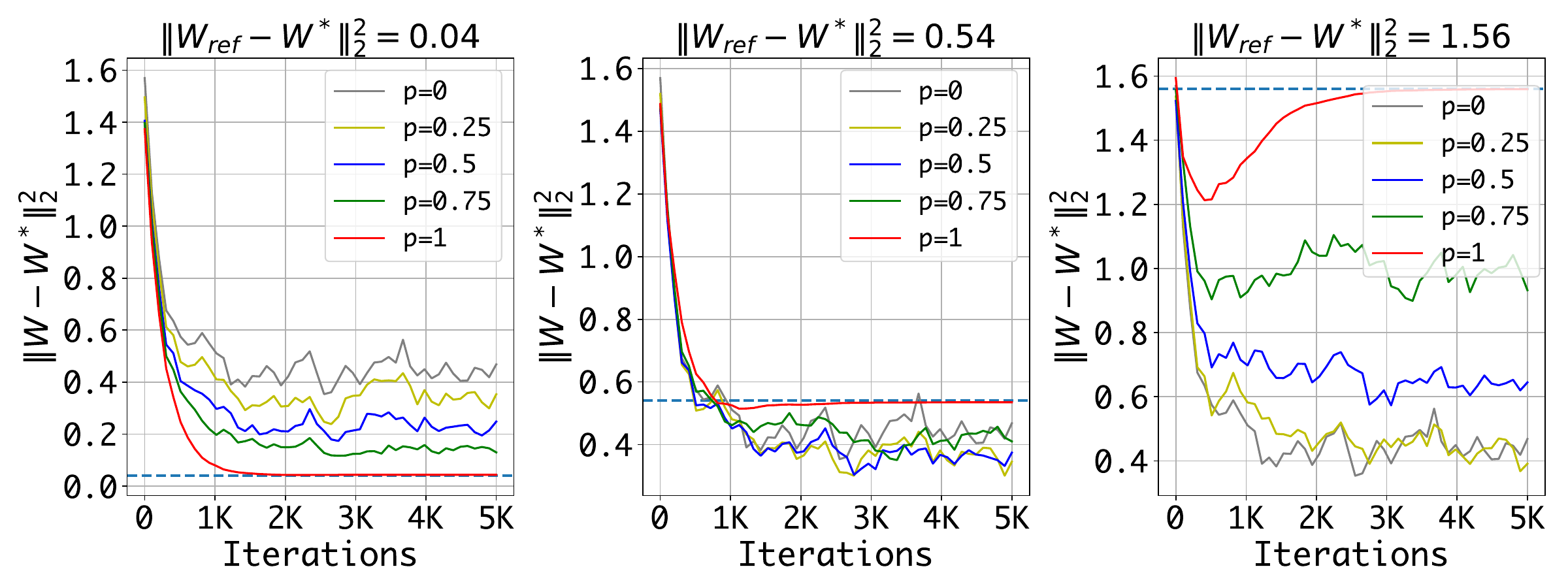}
   % \vspace{-0.8cm}
   \caption{A toy linear regression case. There are real data with noisy labels and synthetic data generated by a reference model $W_{ref}$. The proportion of synthetic data is $p$.}
   \label{fig:2}
   % \vspace{-0.6cm}
\end{figure}

The proof can be found in Appendix~\ref{app:1}. 
Intuitively, Theorem~\ref{thm:1} indicates that, when training data are drawn from a mixture of real and synthetic data points, the distance to the optimal parameters at convergence reflects a trade-off—governed by the synthetic portion $p$—between two factors: the error introduced by label noise in the real data distribution and the discrepancy between the reference model used for generating synthetic data and the optimal model, shown in the 2nd and 3rd terms of Eq.~\ref{eq:3} separately. 
Fig.~\ref{fig:2} provides an illustrative example to visualize this effect. 

Closely examining Eq.\ref{eq:3}, we can discover that, by diminishing label noise, synthetic data would be helpful if the reference model providing these data is accurate. 
Although some works reveal that synthetic data can result in model collapse~\cite{dohmatob2024strong,dohmatob2024model} by amplifying the gap between real and synthetic distributions, we demonstrate that they are useful particularly for ultra-resolution adaption. 
On one hand, large-scale real datasets such as LAION-5B~\cite{schuhmann2022laion} tend to be noisy, containing numerous low-quality images and mismatched text-image pairs. 
On the other hand, at 2K resolution, models like FLUX-1.1 [Pro] Ultra—although closed-weight—are available to produce high-quality synthetic data. 
Building on this analysis, we train our 2K-generation model using only synthetic data in this work, demonstrating superior performance across various scenarios. 

\subsection{Tune Major or Minor Components?}

Parameter-efficient fine-tuning strategies enable adapting a pre-trained model from its original domain to a target domain by integrating lightweight adapters. 
For instance, in personalized text-to-image generation such as DreamBooth~\cite{ruiz2023dreambooth}, attaching low-rank, e.g., rank $r=4$, adapters, \textit{i.e.}, LoRA, to the original model's weights can achieve satisfactory performance~\cite{hu2022lora}. 
Specifically, this is achieved by:
\begin{equation}
    Y=XW+XAB,\label{eq:4}
\end{equation}
where $X$, $Y$, and $W$ are input, output, and original weight matrices respectively,  $A\in\mathbb{R}^{c_{in}\times r}$ and $B\in\mathbb{R}^{r\times c_{out}}$ are low-rank matrices for adaptation, and $c_{in}$ and $c_{out}$ are input and output dimensions, output dimension separately. 
In practice, $A$ is initialized using a normal distribution, whereas $B$ is set to all zeros, which makes the adapter branch output zero initially and allows tuning to begin from the original parameters. 
After tuning, $A$ and $B$ can be merged into the original weight matrix via $W'=W+AB$, ensuring that the total number of model parameters remains unchanged. 
These low-rank adapters employ a small number of parameters that focus on the major components with the largest singular values, enabling efficient adaptation to the target domain~\cite{meng2024pissa}. 

However, different from DreamBooth modifying the styles and appearances of output images, ultra-resolution adaptation focuses on learning the arrangements of details and local textures, which may not correspond to the major components in weight matrices. 
Under this hypothesis, we introduce a method to tune the components associated with the smallest singular values instead. 

Specifically, given a weight matrix $W\in\mathbb{R}^{c_{in}\times c_{out}}$ and $c=\min (c_{in},c_{out})$, we first conduct Singular Value Decomposition (SVD) and derive $W=U\Sigma V$, where $U\in\mathbb{R}^{c_{in}\times c}$ and $V\in\mathbb{R}^{c\times c_{out}}$ are orthogonal matrices, and $\Sigma\in\mathbb{R}^{c\times c}$ is a diagonal matrix with the singular values arranged from large to small. 
Then, $r$ components with the smallest singular values and the rest $c-r$ ones are extracted via:
\begin{equation}
\begin{aligned}
    W^{small}=U[:,-r:]\Sigma[-r:,-r:]V[-r:,:],\\
    W^{res}=U[:,:-r]\Sigma[:-r,:-r]V[:-r,:],\label{eq:5}
\end{aligned}
\end{equation}
where the indexing syntax in Numpy~\cite{harris2020array} and PyTorch~\cite{paszke2019pytorch} are used to represent the operations for extracting multiple rows/columns. 
Analyzing from Eq.~\ref{eq:5}, $W^{small}$ is a low-rank matrix. 
Therefore, for parameter efficiency, we formulate the training time behavior similar to Eq.~\ref{eq:4}:
\begin{equation}
    \begin{aligned}
        &Y=XW^{res}+XAB,\\
        A&=U[:,-r:]\sqrt{\Sigma[-r:,-r:]},\\
        B&=\sqrt{\Sigma[-r:,-r:]}V[-r:,:].\label{eq:6}
    \end{aligned}
\end{equation}
$A$ and $B$ are initialized using Eq.~\ref{eq:6} and updated during fine-tuning. 
In terms of formulation, the approach is similar to PISSA~\cite{meng2024pissa}; however, it fundamentally differs by tuning the components with the smallest singular values instead of the largest. 
Although \cite{wang2024milora} introduce a similar approach in the field of large language model finetuning, they fail to analyze its applicability in various scenarios. 

Empirically, we observe that this approach is particularly effective when no synthetic data are available to serve as a reliable reference, \textit{e.g.}, in 4K generation. 
We speculate that the effectiveness stems from preserving the major components in the original weight matrices, thereby safeguarding the model’s capacity to handle semantics, layouts, and appearances from label noise in the real data distribution. 

\subsection{Enable or Disable Classifier-Free Guidance?}

\begin{figure}[t]
  \centering
   \includegraphics[width=\linewidth]{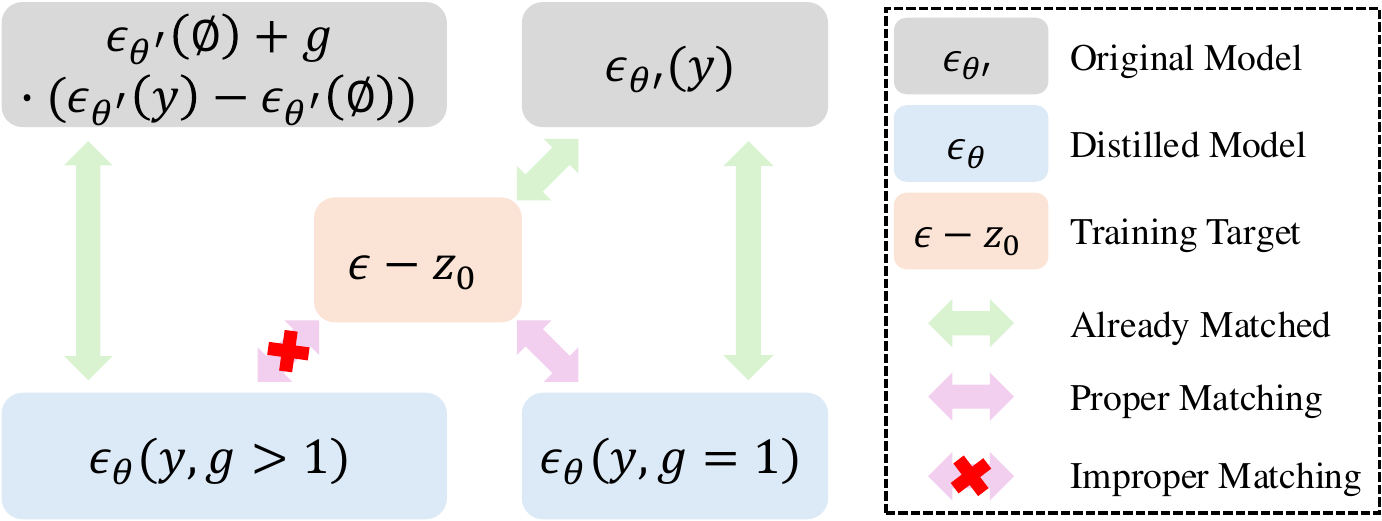}
   % \vspace{-0.8cm}
   \caption{For CFG-distilled models, classifier-free guidance should be disabled in the training time. $z_t$ and $t$ are omitted from the inputs of $\epsilon_{\theta}$ and $\epsilon_{\theta'}$ here for simplicity.}
   \label{fig:3}
   \vspace{-0.4cm}
\end{figure}

Classifier-free guidance (CFG)~\cite{ho2022classifier} aims to enhance the quality of the generated samples by introducing an additional ``null-condition" branch. 
Specifically, at each denoising step in the inference time, the current latent map is processed by both the main branch and the null-condition branch, and the final output is then guided, with a certain strength, in the direction opposing the null-condition branch’s prediction:
\begin{equation}
    \epsilon_{\theta}(z_t,t,\emptyset)+g\cdot(\epsilon_{\theta}(z_t,t,y)-\epsilon_{\theta}(z_t,t,\emptyset)),\label{eq:7}
\end{equation}
where $\emptyset$ denotes the null condition, \textit{e.g.}, an empty prompt, and $g$ is a hyper-parameter controlling the strength. 

\begin{table*}[ht]
\caption{Quantitative results of the baseline methods and our proposed guidelines. The prompts are from HPD and DPG datasets. All images are at a resolution of 2048 $\times$ 2048. Here, FLUX-1.dev$^*$ is FLUX-1.dev with scaled RoPE, proportional attention, and removing dynamic shifting strategies.}
\label{tab:baseline}
\vspace{-0.4cm}
\begin{center}  
\resizebox{\linewidth}{!}{
\begin{tabular}{cccc|cc|cccccccc}
\toprule
\midrule
\multirow{2}{*}{\bf Method/Setting} && \multicolumn{6}{c}{\textbf{HPD Prompt}} && \multicolumn{1}{c}{\textbf{DPG Prompt}} &  \\
\cmidrule{3-8} \cmidrule{10-11}
 && FID ($\downarrow$) & LPIPS ($\downarrow$) & MAN-IQA ($\uparrow$) & QualiCLIP ($\uparrow$) & HPSv2.1 ($\uparrow$) & PickScore ($\uparrow$) && DPG Bench ($\uparrow$)\\
\midrule
\midrule
FLUX1.1 [Pro] Ultra && - & - & 0.4129 & 0.6424 & 29.61 & 22.99 && 84.76\\
\midrule
\midrule
Real-ESRGAN && 36.25 & 0.6593 &  0.4653 & 0.6392  & 30.70 & 22.91 && 83.50 \\
SinSR && 35.09 & 0.6566 &  0.4194 & 0.5556 & 30.95 & 22.96 && 83.79 \\
\midrule
SDEdit && 35.59 & 0.6456 &  0.3736 & 0.4480 & 30.92 & 22.86 && 83.56 \\
\rowcolor{blue!20}
w/ URAE && \textbf{34.07} & \textbf{0.6419} &  \textbf{0.3872} & \textbf{0.5800} & \textbf{32.26} & \textbf{23.02} && \textbf{84.61}\\
I-Max && 33.66 & 0.6394 &  0.3670 & 0.4797 & 31.12 & 23.02 && 83.92\\
\rowcolor{blue!20}
w/ URAE && \textbf{32.24} & \textbf{0.6357} &  \textbf{0.3833} &\textbf{ 0.5736} & \textbf{32.37} & \textbf{23.18} && \textbf{87.88}\\
\midrule
\midrule
PixArt-Sigma-XL && 36.58 & 0.6801 &  0.2949 & 0.4438 & 30.66 & 22.92 && 80.60\\
Sana-1.6B && 33.17 & 0.6792 &  0.3695 & 0.6718 & 30.92 & 22.83 && 85.14\\
\midrule
FLUX-1.dev && 43.78 & 0.6530 & 0.3821 & 0.3800 & 26.22 & 21.54 && 80.64\\
FLUX-1.dev$^*$ && 34.86 & 0.6036 &  0.4110 & 0.5468 & 28.73 & 22.68 && 80.15\\
\rowcolor{blue!20}
w/ URAE && \textbf{29.44} & \textbf{0.5965} & \textbf{0.4730} & \textbf{0.7191} & \textbf{31.15} & \textbf{23.15} && \textbf{83.83}\\
\midrule
\bottomrule
\end{tabular}
}
\end{center}
\vspace{-0.4cm}
\end{table*}

Although effective, the additional null-condition branch doubles the inference cost. 
To address this issue, models like FLUX.1-dev use guidance distillation to train a distilled model that takes the CFG scale embedding as an additional input, encouraging its output aligns with the result in Eq.~\ref{eq:7}. 
Since $g$ is typically larger than $1$ in inference, in training, many works also set $g$ to the same value used at inference time during fine-tuning~\cite{xflux,fluxkits}. 
However, according to the experiments, it results in inferior performance, especially in the problem of ultra-resolution adaptation.

\begin{figure}[t]
  \centering
   \includegraphics[width=0.85\linewidth]{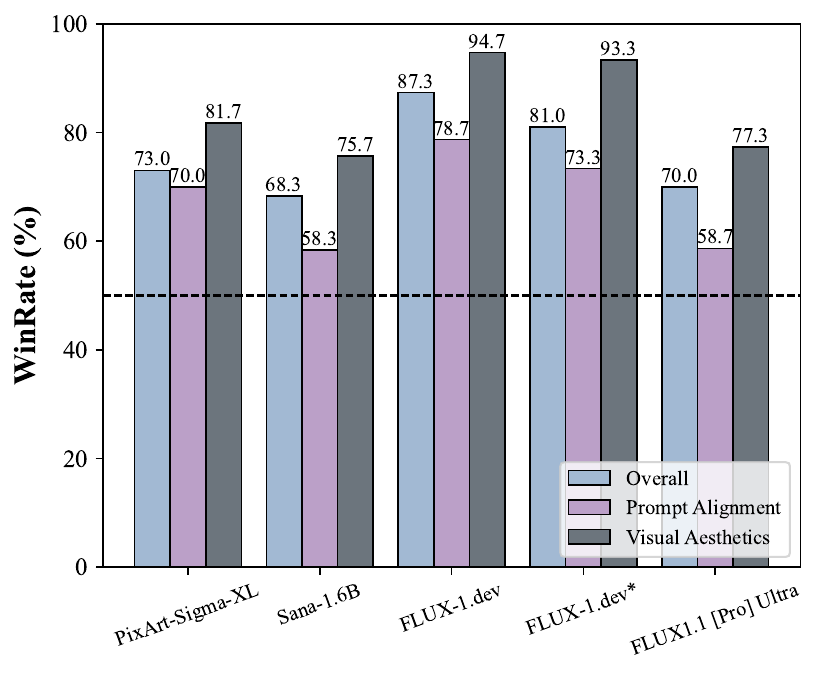}
   \vspace{-0.3cm}
   \caption{GPT-4o preferred evaluation against current SOTA T2I models. We request GPT-4o to select a better image regarding overall quality, prompt alignment, and visual aesthetics. Our proposed method are preferred against others.}
   \vspace{-0.7cm}
   \label{fig:gpt}
\end{figure}

\begin{figure*}[t]
  \centering
   \includegraphics[width=\linewidth]{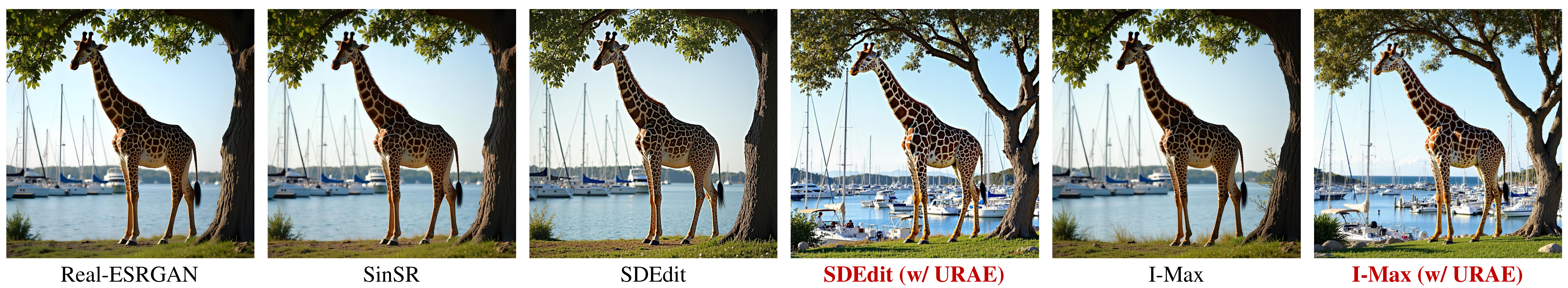}
   \vspace{-0.5cm}
   \caption{Visualizations of our proposed method apply to training-free high-resolution generation pipelines. The prompt is \textit{A giraffe stands beneath a tree beside a marina.}}
   \label{fig:effect-pipe}
   \vspace{-0.5cm}
\end{figure*}

\begin{figure*}[t]
  \centering
   \includegraphics[width=\linewidth]{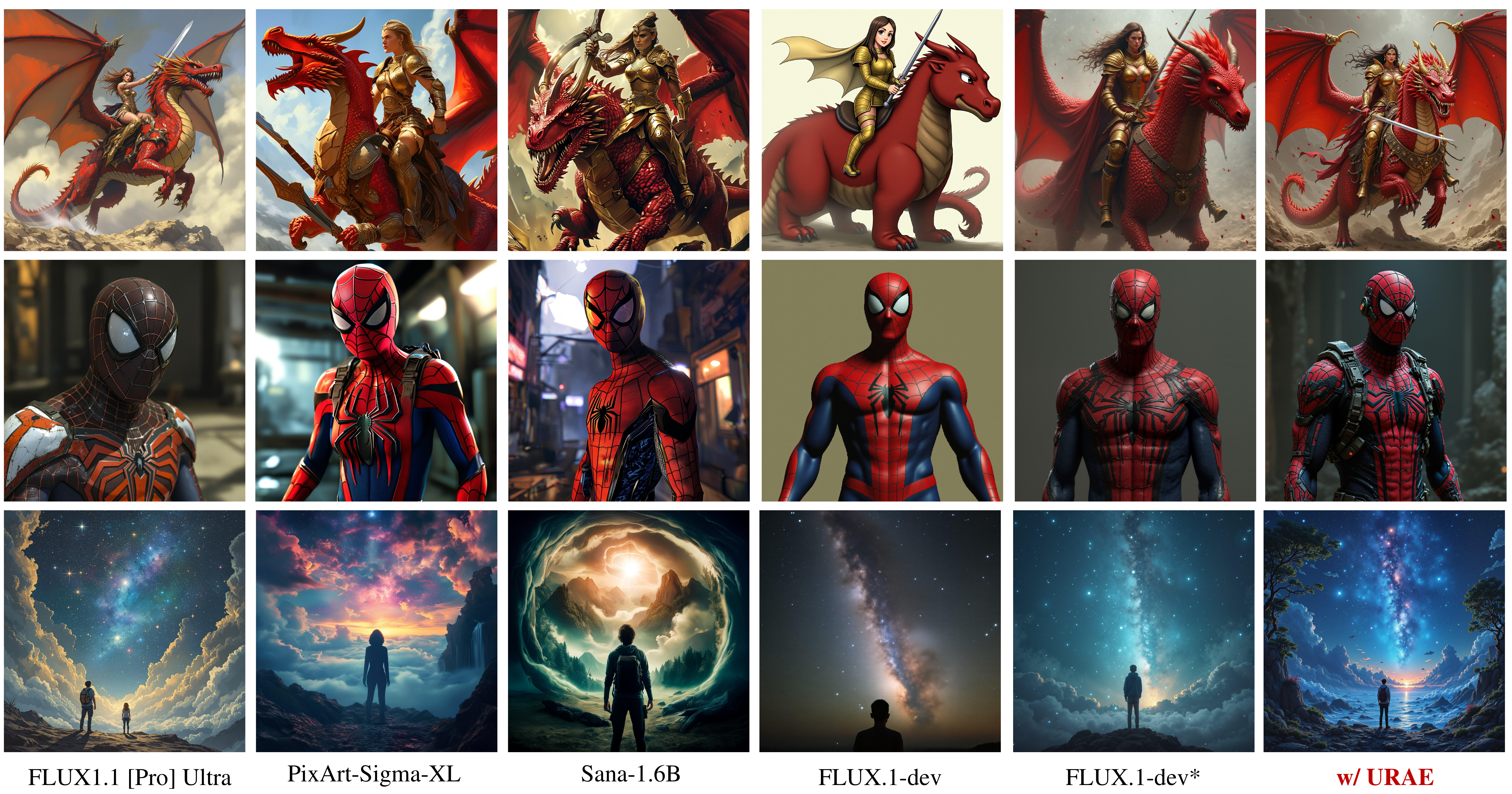}
   \vspace{-0.6cm}
   \caption{Qualitative comparisons with baseline methods. All the images are of 2048 $\times$ 2048 size.}
   \label{fig:baseline}
   \vspace{-0.5cm}
\end{figure*}

% \begin{table}[ht]
% \begin{center}  
% \caption{Results on human preference study. Evaluation images are generated with HPD prompts with a resolution of 2048 $\times$ 2048.}
% \label{tab:human}
% \resizebox{\linewidth}{!}{
% \begin{tabular}{ccc | cc}
% \toprule
% \midrule
% \bf Method/Setting & HPSv2.1 ($\uparrow$) & Rank & PickScore ($\uparrow$) & Rank \\
% \midrule
% \midrule
% SDEdit & 30.92 & 2 & 22.86 & 2\\
% \rowcolor{blue!20}
% w/ URAE & 32.26 & 1 & 23.02 & 1\\
% \midrule
% I-Max & 31.12 & 2 & 23.02 & 2\\
% \rowcolor{blue!20}
% w/ URAE & 32.37 & 1 & 23.18 & 1\\
% \midrule
% \midrule
% PixArt-Sigma-XL & 30.66 & 3 & 22.92 & 3\\
% Sana-1.6B & 30.92 & 2 & 22.83 & 4\\
% FLUX1.1 [Pro] Ultra & 29.61 & 4 & 22.99 & 2\\
% FLUX-1.dev & 26.22 & 6 & 21.54 & 6\\
% FLUX-1.dev$^*$ & 28.73 & 5 & 22.68 & 5\\
% \rowcolor{blue!20}
% w/ URAE & 31.15 & 1 & 23.15 & 1\\
% \midrule
% \bottomrule
% \end{tabular}
% }
% \end{center}
% \end{table}

Specifically, during the distillation stage, the distilled model is trained with $g>1$ as input Eq.~\ref{eq:7}, which involves the null condition. 
In contrast, during the adaptation stage, the target is $\epsilon-z_0$ defined in Eq.~\ref{eq:1}, which is irrelevant to the null condition. 
If $g$ remains larger than $1$, a mismatch arises between the training targets in these two stages, making the training process more challenging. 

To address this issue, we note that incorporating the null-condition branch during adaptation is unnecessary; simply disabling CFG at training time by setting $g=1$ works well and yields a consistent target across the two stages. 
Fig.~\ref{fig:3} illustrates the mismatch triggered by $g>1$ and how $g=1$ addresses the problem. 

During inference, CFG is still necessary by using $g>1$. 
Although the model does not encounter $g>1$ during adaptation, we find that it generalizes sufficiently well in practice.

\section{Experiments}

\subsection{Settings and Implementation Details}
In this paper, we adopt the open-source text-to-image FLUX.1-dev model~\cite{flux2023} as the base model to demonstrate the effectiveness of our proposed URAE guidelines, thanks to its superior performance. For our 2K-generation model, we collect 3K synthetic samples with various aspect ratios generated by the FLUX1.1 [Pro] Ultra model as the training dataset, and fine-tune the FLUX.1-dev on it for merely 2K iterations with a batch size of 8, which takes only $\sim1$ day on 2 H100 GPUs. 
For our 4K model, we utilize 30K images with at least 4K resolution from the LAION-5B dataset~\cite{schuhmann2022laion} and fine-tune the base model FLUX.1-dev for 2K iterations on 8 H100 GPUs, which takes $\sim1$ days. 
In terms of training convergence, our method requires significantly fewer iterations compared with state-of-the-art methods, such as 10K for SANA~\cite{xie2024sana}. 

For baseline, we apply URAE on the FLUX.1-dev model and compare the performance with PixArt-Sigma-XL~\cite{chen2024pixartsigma}, Sana-1.6B~\cite{xie2024sana}, and FLUX series models. 
% , including FLUX1.1 [Pro] Ultra, which are capable of generating high resolution images directly. 
In order to further demonstrate the effectiveness of URAE, we also apply URAE to the existing training-free high-resolution generation pipelines, \textit{i.e.}, SDEdit~\cite{meng2021sdedit} and I-Max~\cite{du2024max}. These pipelines require the base text-to-image model, \textit{e.g.}, FLUX.1-dev, to generate low-resolution, \textit{e.g.}, 1024 $\times$ 1024, images as the guidance, and upscale these images to higher resolutions through image-to-image pipelines. For comparison, we also include the super-resolution methods Real-ESRGAN~\cite{wang2021real} and SinSR~\cite{wang2024sinsr}, based on GAN and diffusion, respectively. 
For baseline comparison, we conduct quantitative experiments on 2048 $\times$ 2048 samples generated with prompts from the HPD~\cite{wu2023human} and DPG~\cite{hu2024ella} datasets.
% For baseline comparison, we use FID and STLPIPS to evaluate perceptual quality and similarity with the reference dataset generated by the FLUX1.1 [Pro] Ultra model. Furthermore, the quality of the generated images is assessed using metrics such as MUSIQ, CLIPIQA, MANIQA, and Q-Align. We adopt the DPG Bench to measure the semantic consistency and coherence between the generated image and the corresponding prompt.
Additionally, we use the HPSv2.1~\cite{wu2023human} and PickScore~\cite{kirstain2023pick} as human preference metrics to further evaluate the quality and aesthetic appeal of the generated images. Following prior works like PixArt-Sigma and I-Max, we also utilize the GPT-4o to assess the generated images from three key perspectives: prompt alignment, visual aesthetics, and overall quality, at both 2K and 4K resolutions. These AI preference scores are derived from 300 randomly selected prompts in the COCO30K~\cite{lin2014microsoft,chen2024pixartsigma} dataset. 

\begin{figure*}[t]
  \centering
   \includegraphics[width=\linewidth]{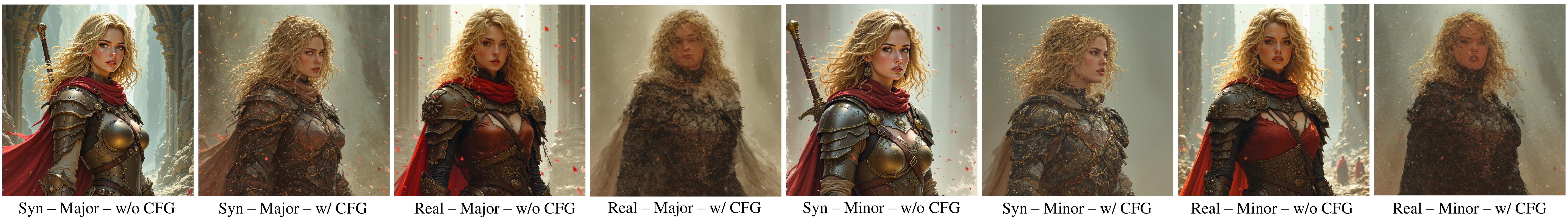}
   \vspace{-0.6cm}
   \caption{Visualization results of ablation studies. The prompt is \textit{Imogen Poots portrayed as a D\&D Paladin in a fantasy concept art by Tomer Hanuka}.}
   \label{fig:vis-abl}
   \vspace{-0.4cm}
\end{figure*}

\begin{figure*}[t]
  \centering
   \includegraphics[width=\linewidth]{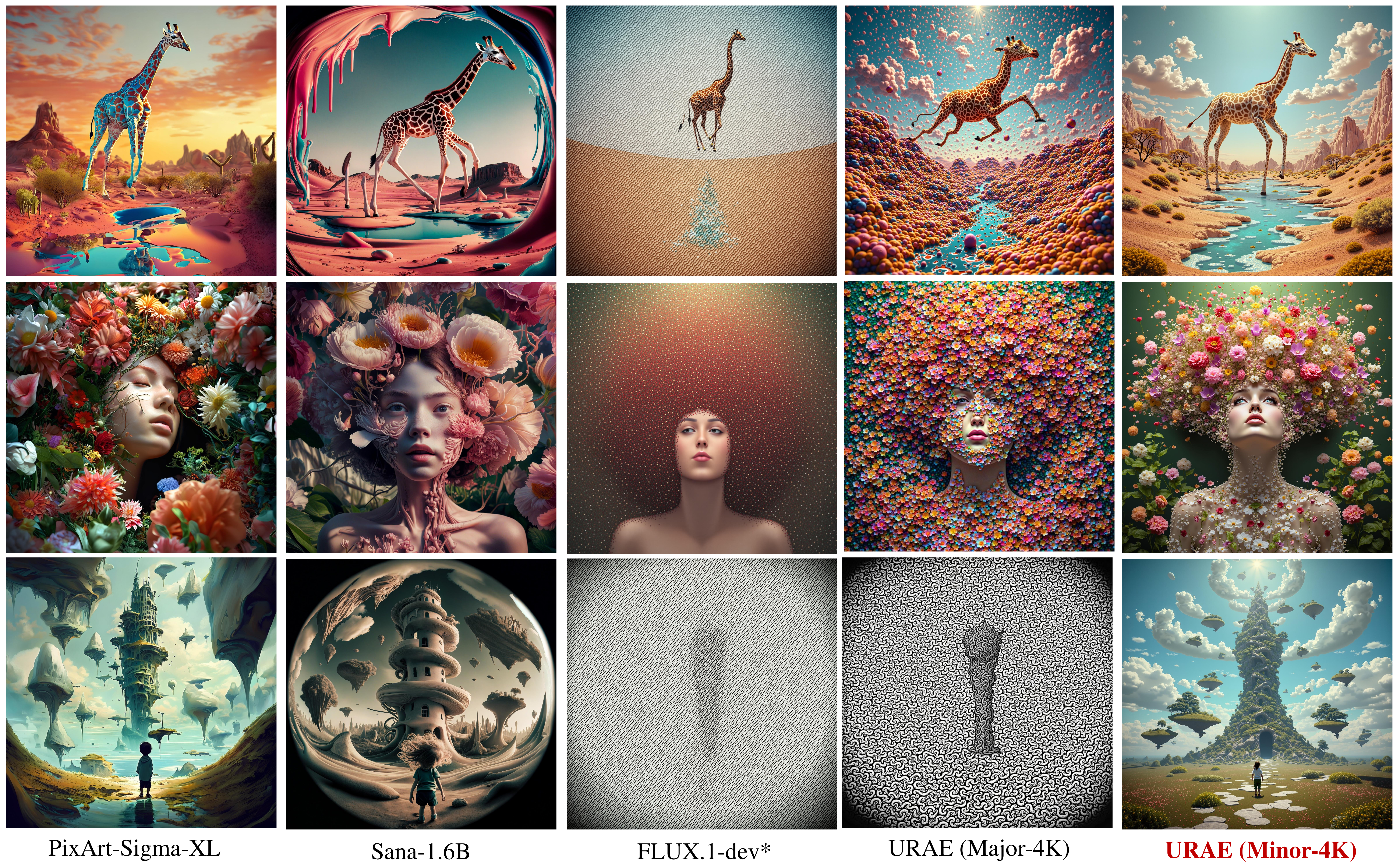}
   \vspace{-0.6cm}
   \caption{Visualization results for ultra-resolution image generation task. All the images are of 4096 $\times$ 4096 size.}
   \label{fig:effect-pipe-4k}
   \vspace{-0.4cm}
\end{figure*}

\subsection{2K Resolution}
Here, we evaluate the performance of the proposed methods on 2K images generated with prompts from HPD and DPG datasets. The results are shown in Table~\ref{tab:baseline}. 
The quantitative results indicate that our proposed method is capable of significantly enhancing the ability of models to generate high-resolution images and demonstrates its versatility and adaptability across different methods. Our method surpasses the state-of-the-art model FLUX1.1 [Pro] Ultra in terms of image quality, demonstrating its superiority in generating visually refined images. Moreover, the remarkable improvements in image quality further underscore the strength of our method in achieving state-of-the-art visual results for all quality metrics, making it a highly effective solution for high-resolution image generation tasks. In addition, our method also achieves a substantial improvement in the performance of the base model in terms of prompt alignment, improving the original FLUX-1.dev by 3.19 in DPG Bench score, and 3.96 for I-Max pipeline. 

% \subsection{Human and AI Preference Study}
For human preference study, we adopt HPSv2.1 and PickScore to benchmark the human preference score. The samples are generated with prompts from the HPD dataset and the resolution is 2048 $\times$ 2048. The results are shown in Table~\ref{tab:baseline}. The results show that our method improves the human preference score of the base model, indicating that our proposed guidelines are capable of generating images that better align with human preferences.

We also conduct AI preference studies with GPT-4o for pair comparison regarding overall quality, prompt alignment, and visual aesthetics aspects. The results are shown in Fig.~\ref{fig:gpt}. For the prompts for GPT-4o to assess the quality, prompt alignment, and visual aesthetics, please refer to Appendix~\ref{ai-prompts}. The results demonstrate that our proposed method excels and is preferred in all three aspects. Please refer to Appendix~\ref{ai-study} for more quantitative results.

\begin{table*}[ht]
\begin{center} 
\caption{Ablation studies on three key guidelines, using real (Real) or synthetic (Syn) data, whether to adopt CFG in training, and tuning of major or minor components. Evaluations are on the 2048 $\times$ 2048 images generated from HPD prompts.}
\label{tab:ablation}
\vspace{-0.2cm}
\resizebox{\linewidth}{!}{
\begin{tabular}{clcc|cc|cccccccc}
\toprule
\midrule
 &\bf Method/Setting & FID ($\downarrow$) & LPIPS ($\downarrow$) &  MAN-IQA ($\uparrow$) & QualiCLIP ($\uparrow$) & HPSv2.1 ($\uparrow$) & PickScore ($\uparrow$) \\
\midrule
\midrule
% \rowcolor{blue!20}
\multirow{4}{*}{\bf Major} 
& Syn w/o CFG& 29.44 & 0.5965 & \textbf{0.4730} & \textbf{0.7191} & \textbf{31.15} & \textbf{23.15}\\
&Syn w/ CFG &76.07 & 0.6388 & 0.3992 & 0.5890 & 24.80 & 21.87\\
&Real w/o CFG & 31.39 & 0.6076 & 0.4262 & 0.5953 & 29.33 & 22.86\\
&Real w/ CFG & 133.68 & 0.5978 & 0.3254 & 0.3645 & 16.55 & 19.92\\
\midrule
\multirow{4}{*}{\bf Minor} 
&Syn w/o CFG & \textbf{27.90} & \textbf{0.5779} &  0.4558 & 0.6616 & 30.40 & 22.87\\
&Syn w/ CFG & 65.34 & 0.5858 &  0.3852 & 0.5312 & 23.77 & 21.64\\
&Real w/o CFG & 32.09 & 0.6000 &  0.4485 & 0.6098 & 28.71 & 22.61\\
&Real w/ CFG & 133.32 & 0.6026 &  0.3387 & 0.3672 & 16.36 & 19.68\\

\midrule
\bottomrule
\end{tabular}
}
\end{center}
\vspace{-0.4cm}
\end{table*}

\begin{table*}[ht]
\begin{center} 
\caption{Evaluation on ultra resolution image generation task. The images are of 4096 $\times$ 4096, and generated with prompts randomly selected from COCO30K. For user study, the prompts are randomly generated as listed in Appendix~\ref{user}.}
\vspace{-0.2cm}
\label{tab:ultra-res}
\resizebox{\linewidth}{!}{
\begin{tabular}{ccccccccccccccccc}
\toprule
\midrule
\multirow{2}{*}{\bf Method/Setting} && \multicolumn{4}{c}{\textbf{HPD Prompt}} && \multicolumn{6}{c}{\textbf{User Study}} &  \\
\cmidrule{3-6} \cmidrule{8-14}
 && MAN-IQA ($\uparrow$) & Rank & QualiCLIP ($\uparrow$) & Rank && Overall Quality & Rank & Prompt Alignment & Rank & Visual Aesthetic & Rank \\
\midrule
\midrule
PixArt-Sigma-XL && 0.2935 & 5 & 0.2308 & 5 && 31.18\% & 2 & 30.88\% & 2 & 30.00\% & 2 \\
Sana-1.6B && 0.3288 & 3 & 0.4979 & 2 && 10.29\% & 3 & 10.00\% & 3 & 12.06\% & 3 \\
FLUX-1.dev$^*$ && 0.3673 & 2 & 0.2564 & 4 && 3.24\% & 4 & 3.24\% & 4 & 3.24\% & 4 \\
w/ URAE~(Major-4K) && 0.3280 & 4 & 0.2700 & 3 && 2.06\% & 5 & 1.76\% & 5 & 1.76\% & 5 \\
w/ URAE~(Minor-4K) && \textbf{0.3999} & 1 & \textbf{0.5118} & 1 && \textbf{53.24\%} & 1 & \textbf{54.12\%} & 1 & \textbf{52.94\%} & 1 \\
\midrule
\bottomrule
\end{tabular}
}
\end{center}
\vspace{-0.5cm}
\end{table*}

\subsection{4K Resolution}
Here, we evaluate the performance of our proposed method in 4K-ultra-resolution image generation. 
The results are shown in Table~\ref{tab:ultra-res}. 
From the experimental results, fine-tuning minor components achieves outstanding performance when no synthetic data are available to serve as a reliable reference for the ultra-resolution image generation task, while the commonly adopted LoRA may fail on the overall semantics. Moreover, our proposed method also demonstrates exceptional performance compared with other methods, further validating its competitiveness to existing approaches.

Given that there is no well-defined benchmark specifically for 4K-ultra-resolution generation, we conduct a user study using randomly generated prompts listed in Appendix~\ref{user}. 
For each prompt, we present results generated by the candidates to users and let them select the best one from three aspects including overall quality, prompt alignment, and visual aesthetic aspects. 
We collect $1,020$ votes altogether. 
The results in Table~\ref{tab:ultra-res} are coherent with the above analysis.

\subsection{Ablation Study}
Here, to evaluate the effectiveness of our proposed method, we carry out ablation studies on three key guidelines that we proposed in URAE, the source of training data, tuning major or minor components, and whether to adopt classifier-free guidance (CFG) in training. The experiments are conducted on the 2048 $\times$ 2048 images generated from HPD prompts. The results are shown in Table~\ref{tab:ablation}, and the visualization examples are shown in Fig.~\ref{fig:vis-abl}. For the source of training data, the results demonstrate that high-quality synthetic data can provide better performance than noisy real data. When the model is fine-tuned with real data, tuning minor components can bring more vivid details as shown in Fig.~\ref{fig:vis-abl}. As for CFG, although it is necessary in the inference stage, it can lead to significant performance degradation in the training stage. 
According to these results, we by default use synthetic data and tune major components, \textit{i.e.} adopt LoRA, at 2K resolution, while use real data and tune minor components at 4K resolution. 
In both cases, CFG is disabled in training and enabled during inference.

\section{Conclusions and Limitations}
In this paper, we focus on the challenge of adapting text-to-image diffusion models from their native scales to ultra-resolution settings with limited training data and computational resources. 
Our proposed framework, URAE, tackles the problem from two complementary perspectives, \textit{i.e.}, data and parameter efficiency, and provides a set of useful guidelines. 
First, by incorporating synthetic data generated by some teacher models, we demonstrate the potential to promote training convergence and achieve high-quality outcomes even under data-scarce conditions. 
Second, for the cases where synthetic data are unavailable, we introduce a parameter-efficient fine-tuning approach to tune the minor components of weight matrices, which outperforms standard low-rank adapters. 
Additionally, for models employing guidance distillation, e.g., FLUX, setting the guidance scale to $1$ during adaptation proves crucial for achieving favorable results. 
Extensive experiments reveal that URAE matches the 2K-generation performance of leading closed-source solutions such as FLUX1.1 [Pro] Ultra using only 3K samples and 2K iterations, and further sets new milestones for 4K-resolution generation. 

Nevertheless, the models presented in this work fall short of matching the inference-time efficiency exhibited by recent high-resolution text-to-image generation methods~\cite{xie2024sana,liu2024linfusion,liu2024clear,chen2024pixartsigma}, as we have not introduced architectural optimizations specifically targeting this aspect. 
In the future, we envision research aimed at streamlining the ultra-resolution generation process to balance quality and efficiency requirements in practice. 
It is also meaningful to integrate our methods into multi-modal large language models to unlock even broader and more versatile capabilities. 

\section*{Impact Statement}

Our work on ultra-resolution text-to-image generation has broad implications for both research and real-world applications. By pushing resolution boundaries, we enable richer and more detailed visual content, benefiting domains such as digital art, virtual reality, advertising, and scientific visualization. At the same time, these advancements highlight important ethical considerations. High-fidelity images could be misused for creating deceptive content, and the computational demands of large-scale generation can have environmental impacts. We therefore emphasize responsible development practices, including efficient training strategies and transparent model documentation, to help ensure that the benefits of ultra-resolution text-to-image models are realized while mitigating potential risks. 

\bibliography{example_paper}
\bibliographystyle{icml2025}

%%%%%%%%%%%%%%%%%%%%%%%%%%%%%%%%%%%%%%%%%%%%%%%%%%%%%%%%%%%%%%%%%%%%%%%%%%%%%%%
%%%%%%%%%%%%%%%%%%%%%%%%%%%%%%%%%%%%%%%%%%%%%%%%%%%%%%%%%%%%%%%%%%%%%%%%%%%%%%%
% APPENDIX
%%%%%%%%%%%%%%%%%%%%%%%%%%%%%%%%%%%%%%%%%%%%%%%%%%%%%%%%%%%%%%%%%%%%%%%%%%%%%%%
%%%%%%%%%%%%%%%%%%%%%%%%%%%%%%%%%%%%%%%%%%%%%%%%%%%%%%%%%%%%%%%%%%%%%%%%%%%%%%%
\newpage
\appendix
\onecolumn

\section{Related Works}

In this section, we briefly introduce related works and their relationships between this paper from three aspects: text-to-image diffusion models, high-resolution generation, and parameter-efficient fine-tuning. 

\subsection{Text-to-Image Diffusion Models}

The diffusion model~\cite{ho2020denoising} has emerged as a powerful class of generative models. 
Unlike traditional approaches such as GANs~\cite{goodfellow2014generative}, diffusion models iteratively refine noisy maps with a UNet backbone~\cite{ronneberger2015u} to produce high-quality and detailed images~\cite{nichol2021improved,dhariwal2021diffusion}, which fuels significant advancements in large-scale text-to-image diffusion models~\cite{rombach2022high,podell2023sdxl,balaji2022ediffi,ding2022cogview2,nichol2021glide,ramesh2022hierarchical,razzhigaev2023kandinsky,xu2023versatile,saharia2022photorealistic}. 
Leveraging billions of image-text pairs, they demonstrate remarkable semantic understanding and the ability to generate diverse and photorealistic images aligning with text prompts. 

Most recently, Transformer~\cite{vaswani2017attention} has been introduced as an alternative backbone to UNet~\cite{Peebles2022DiT} in diffusion models, known as Diffusion Transformer (DiT). 
Then, text-to-image models based on it have progressively demonstrated dominant performance~\cite{chen2024pixartsigma,esser2024scaling,gao2024lumina,li2024hunyuan,zheng2024cogview3}. 
We thus focus on DiT-based models and conduct experiments mainly on FLUX, which yields state-of-the-art text-to-image performance, in sake for superior ultra-resolution adaption results. 

\subsection{High-Resolution Generation}

Training models at high resolutions demands substantial computational resources. 
To address this, a series of works propose training-free solutions, developing inference stage strategies that allow diffusion models trained at their native resolutions to operate effectively at higher scales~\cite{bar2023multidiffusion,meng2021sdedit,du2024max,he2023scalecrafter,du2024demofusion,huang2024fouriscale,wu2024megafusion,zhang2023hidiffusion}. 
While effective, without looking at any high-resolution images during training, in fact, they still fall short in accurately handling detailed structures and textures inherent in ultra-resolution images. 
By contrast, ultra-resolution adaptation focused in this paper dedicates on addressing this drawback through training, which is technically orthogonal to training-free approaches and can work as a plug-and-play component to enhance their performance. 

There are indeed some works training for high-resolution generation like 4K~\cite{chen2024pixartsigma,xie2024sana,zheng2024cogview3,ren2024ultrapixel}. 
However, millions of high-quality training data and industrial-scale computational resources are required to train the whole transformer backbone. 
In this paper, we focus on the challenges of data and parameter efficiency and demonstrate that comparable or even superior performance can be achieved with significantly less data and fewer trainable parameters. 

Another line of research has concentrated on enhancing inference efficiency through the development of efficient and scalable diffusion backbones~\cite{chen2024pixartsigma,liu2024linfusion,liu2024clear}. 
These designs and insights are orthogonal to our work, and it is promising to combine their strengths with our approach to achieve the best of both training and inference efficiency, which lies beyond the scope of this paper and is left for future exploration. 

\subsection{Parameter-Efficient Fine-Tuning}

In many real-world scenarios, fine-tuning existing models for specific applications is often necessary. 
However, fine-tuning all parameters can lead to substantial computational overhead, particularly in terms of memory footprint. 
To address this limitation, a series of works propose parameter-efficient fine-tuning strategies~\cite{hu2022lora,hyeon2021fedpara,meng2024pissa,yeh2023navigating,wang2024milora}. 
In this paper, we aim at an effective method specifically tailored for ultra-resolution adaptation.

\section{Theoretical Proof}\label{app:1}

We supplement the proof of our main theoretical results in Theorem~\ref{thm:1} here. 
\begin{framedtheorem}
Under the setting defined in Assumptions~\ref{ass:1}, \ref{ass:2}, and \ref{ass:3}, the error between $W_T$, the parameters after $T$ training iterations, and the optimal $W^*$ is bounded by:
\begin{equation}
    \begin{aligned}
    \mathbb{E}[\Vert W_T - W^*\Vert_2^2]\leq&\mathbb{E}[\Vert(I - \eta M)^T \Delta_0\Vert_2^2]+
    \eta^2 (p(1-p)\mathbb{E}[\delta^2]+(1-p)\sigma^2) \sum_{i=1}^{N} \frac{(1 - (1 - \eta \lambda_i)^T)^2}{\lambda_i}\\&+p^2\Vert W_{ref}-W^*\Vert_2^2.
    \end{aligned}
\end{equation}
where $\Delta_0=W_0-(pW_{ref}+(1-p)W^*)$, $M$ is defined as $\nabla_{W}f(U;W_0)^\top\nabla_{W}f(U;W_0)$, $\delta=f(u;W_{ref})-f(u;W^*)$, and $\lambda_i$ is the $i$-th eigenvalue of $M$. 
\end{framedtheorem}
\begin{proof}
Under the assumption of infinite-width neural networks, the network output $f(u;W)$ can be viewed as the following linear form with respect to the parameter $W$:
\begin{equation}
    f(u;W)\approx f(u;W_0)+\nabla_W f(u;W_0)(W-W_0).\label{eq:app1}
\end{equation}
We denote $\nabla_W f(W_0;u)$ as $\Phi$ for simplicity. 
According to the loss function $\mathcal{L}=\frac{1}{2N}\Vert f(u;W)-x\Vert_2^2$ and Eq.~\ref{eq:app1}, the gradient of $\mathcal{L}$ with respect to $W$ is:
\begin{equation}
    \nabla_W\mathcal{L}=\frac{1}{N}\sum_{N}\{\Phi^\top(f(u;W)-x)\}=\frac{1}{N}\sum_{N}\{\Phi^\top[\Phi(W-W_0)+f(u;W_0)-x)]\}.
\end{equation}
Training is conducted using SGD:
\begin{equation}
    W_{t+1}=W_{t}-\eta\nabla_W\mathcal{L}_t=W_t-\eta\frac{1}{N}\sum_{N}\{\Phi^\top[\Phi(W_t-W_0)+f(u;W_0)-x)]\}.\label{eq:app3}
\end{equation}
Due to the linearity, the optimal parameter $W^{*'}$ when training on a mixture of real and synthetic data is given by:
\begin{equation}
    W^{*'}=pW_{ref}+(1-p)W^*.\label{eq:app_param}
\end{equation}
The target $x$ can be viewed as the output of $f(u;W^{*'})$ with a noise term $\xi'$:
\begin{equation}
    x=f(u;W^{*'})+\xi'.\label{eq:app4}
\end{equation}
Then we analyze the mean and variance of $\Sigma_{\xi'}$. 
According to Eq.~\ref{eq:app4}, $\xi'$ satisfies:
\begin{equation}
    \xi' =
    \begin{cases}
        f(u;W_{ref})-f(u;W^{*'}), & \text{with probability } p, \\
        f(u; W^*) + \xi - f(u;W^{*'}), & \text{with probability } 1 - p.
    \end{cases}
\end{equation}
Given the linearity and Eq.~\ref{eq:app_param}, 
\begin{equation}
    f(u;W^{*'})=pf(u;W_{ref})+(1-p)f(u;W^*). 
\end{equation}
Denote $f(u;W_{ref})-f(u;W^*)$ as $\delta$. 
Then,
\begin{equation}
    \xi' =
    \begin{cases}
        (1 - p) \delta, & \text{with probability } p, \\
        - p \delta + \xi, & \text{with probability } 1 - p.
    \end{cases}
\end{equation}
Since $\xi\sim\mathcal{N}(0,\sigma^2)$, the expectation of $\xi'$ is:
\begin{equation}
    \mathbb{E}[\xi'] = p(1 - p) \delta + (1 - p)(-p \delta) = 0.
\end{equation}
And the variance, denoted as $\Sigma_{\xi'}$, is computed as:
\begin{equation}
    \Sigma_{\xi'}=\mathbb{E}[\xi'^2] = p(1-p)^2\mathbb{E}[\delta^2]+(1-p)\mathbb{E}[(-p\delta+\xi)^2]=p(1-p)\mathbb{E}[\delta^2]+(1-p)\sigma^2.\label{eq:app_var}
\end{equation}

Let $M$ denote $\Phi^\top\Phi\in\mathbb{R}^{D\times D}$, where $D$ is the number of parameters in the network, and $\Delta_t$ denote $W_t-W^{*'}$. 
Combining with Eqs.~\ref{eq:app3} and \ref{eq:app4}, we obtain:
\begin{equation}
    W_{t+1}=(W^{*'}+\Delta_t)-\eta\frac{1}{N}\sum_{N}\{\Phi^\top[\Phi(\Delta_t+(W^*-W_0))+f(u;W_0)-(f(u;W^{*'})+\xi')]\}.
\end{equation}
Note that:
\begin{equation}
    f(u;W^{*'})-f(u;W_0)\approx\Phi(W^{*'}-W_0).
\end{equation}
We then have:
\begin{equation}
    W_{t+1}=(W^{*'}+\Delta_t)-\eta\Phi^\top[\Phi\Delta_t+\xi']=W^{*'} + [(I-\eta\Phi^\top\Phi)\Delta_t+\eta\Phi\xi'].
\end{equation}
Thus,
\begin{equation}
    \Delta_{t+1}=(I-\eta M)\Delta_t+\eta\Phi\xi'.\label{eq:app5}
\end{equation}
Starting from $\Delta_0$, after $T$ iterations, according to Eq.~\ref{eq:app5}, we can obtain the explicit expression of $\Delta_T$:
\begin{equation}
    \Delta_T = \underbrace{(I - \eta M)^T \Delta_0}_{\text{Initial error decay}} + \underbrace{\eta \sum_{k=0}^{T-1} (I - \eta M)^k \Phi^\top \xi'}_{\text{Label noise accumulation}}.
\end{equation}
Now, we are interested in $\mathbb{E}[\Vert \Delta_T\Vert_2^2]$, which contains quadratic terms of initial error decay and label noise accumulation as well as their cross term. 
Since label noise is independent of error caused by initializing model parameters and $\mathbb{E}[\xi']=0$, the cross term is $0$. 
Thus,
\begin{equation}
    \mathbb{E}[\Vert W_T - W^{*'}\Vert_2^2]
    =
    \mathbb{E}[\Vert(I - \eta M)^T \Delta_0\Vert_2^2]
    +
    \eta^2 \mathbb{E} \Biggl[ \Bigl\Vert \sum_{k=0}^{T-1} (I - \eta M)^k \Phi^\top\xi' \Bigr\Vert_2^2 \Biggr].
\end{equation}
The first term is related to the initial error. 
For the noise term, let \( A_T \) be the cumulative sum of the updates:
\begin{equation}
    A_T = \sum_{k=0}^{T-1} (I - \eta M)^k.
\end{equation}
Then:
\begin{equation}
    \sum_{k=0}^{T-1} \sum_{j=0}^{T-1} (I - \eta M)^k \Phi^\top \xi \xi'^\top \Phi (I - \eta M)^j
    = A_T \Phi^\top \Sigma_{\xi'} \Phi A_T^\top.
\end{equation}
Since the noise covariance $\Sigma_{\xi'}$ is a scalar, using \( M = \Phi^\top \Phi \), we get:
\begin{equation}
    \mathbb{E}[\Vert A_T \Phi^\top \xi' \Vert_2^2]
    = \Sigma_{\xi'} \mathrm{Tr} \big( A_T M A_T^\top \big).
\end{equation}
Using the matrix geometric series sum formula:
\begin{equation}
    A_T = M^\dag (I - (I - \eta M)^T),
\end{equation}
where $^\dag$ denotes the Moore-Penrose pseudoinverse. 
Substituting,
\begin{equation}
    A_T M A_T^\top = M^\dag (I - (I - \eta M)^T) M (I - (I - \eta M)^T) M^\dag.
\end{equation}
Taking the trace:
\begin{equation}
    \mathrm{Tr} \big(A_T M A_T^\top \big) = \mathrm{Tr} \big(M^\dag (I - (I - \eta M)^T)^2 \big).
\end{equation}
Let \( M \) have eigenvalue decomposition:
\begin{equation}
    M = V \Lambda V^\top, \quad \Lambda = \text{diag}(\lambda_1, ..., \lambda_D).
\end{equation}
Note that for infinite-width network, $D\gg N$. 
Thus, $\lambda_i=0$ for $N<i\leq D$. 
Then:
\begin{equation}
    (I - \eta M)^{2T} = V (I - \eta \Lambda)^{2T} V^\top.
\end{equation}
Therefore:
\begin{equation}
    \mathrm{Tr} \big[(I - \eta M)^{2T} \big] = \sum_{i=1}^{N} (1 - \eta \lambda_i)^{2T}.
\end{equation}
Similarly,
\begin{equation}
    \mathrm{Tr} \big[ M^\dag (I - (I - \eta M)^T)^2 \big] = \sum_{i=1}^{N} \frac{(1 - (1 - \eta \lambda_i)^T)^2}{\lambda_i}.
\end{equation}
Thus, using Eq.~\ref{eq:app_var}, the result of $\mathbb{E}[\Vert \Delta_T\Vert_2^2]$ is:
\begin{equation}
    \mathbb{E}[\Vert W_T - W^{*'}\Vert_2^2]
    =
    \mathbb{E}[\Vert(I - \eta M)^T \Delta_0\Vert_2^2]
    + \eta^2 (p(1-p)\mathbb{E}[\delta^2]+(1-p)\sigma^2) \sum_{i=1}^{N} \frac{(1 - (1 - \eta \lambda_i)^T)^2}{\lambda_i}. 
\end{equation}
By triangle inequality, we have:
\begin{equation}
    \begin{aligned}
    \mathbb{E}[\Vert W_T - W^*\Vert_2^2]\leq&\mathbb{E}[\Vert W_T - W^{*'}\Vert_2^2]+\Vert
W^{*'}-W^*\Vert_2^2   \\
    =&\mathbb{E}[\Vert W_T - W^{*'}\Vert_2^2]+\Vert pW_{ref}+(1-p)W^*-W^*\Vert_2^2 \\
    =&\mathbb{E}[\Vert(I - \eta M)^T \Delta_0\Vert_2^2]
    + \eta^2 (p(1-p)\mathbb{E}[\delta^2]+(1-p)\sigma^2) \sum_{i=1}^{N} \frac{(1 - (1 - \eta \lambda_i)^T)^2}{\lambda_i}\\&+p^2\Vert W_{ref}-W^*\Vert_2^2.
    \end{aligned}
\end{equation}
\end{proof}

\section{More Experimental Details}
\subsection{Prompts for AI Preference Study}\label{ai-prompts}
To better compare the quality of generated images, we employ GPT-4o as the evaluator, assessing methods from three aspects: overall quality, visual aesthetics, and prompt alignment. The evaluation involved both pairwise comparisons and quantitative analysis. During the evaluation, for pairwise comparison, GPT-4o compares our method with the baseline methods, selecting the more preferred image. For quantitative analysis, GPT-4o assigns scores (0-100) to each image generated by each method. The prompts used in our testing are listed below, designed following the previous work PixArt-Sigma~\cite{chen2024pixartsigma}.
For pairwise comparison, the designed prompt to evaluate the overall quality of images is as follows:
\begin{tcolorbox}[colframe=gray!70, colback=white, sharp corners]
As an AI visual assistant, you are an evaluator specialized in image quality analysis for high-resolution text-to-image generation models. Given a specific caption, please evaluate the overall quality of the image by considering both content alignment and technical excellence. For content alignment, assess the key information including object identities, properties, spatial relationships, object numbers and caption-specified style. For technical quality, evaluate the image's photorealism and aesthetics, focusing on clarity, richness of detail, artistic quality, and overall visual appeal. Please analyze how well the image performs in both aspects to determine its comprehensive quality, the prompt is \textit{``your prompt"}. Please output [Image 1] if the first image is better, [Image 2] if the second image is better, and give me the reason.
\end{tcolorbox}

\begin{table}[ht]
\begin{center} 
\caption{Results on AI preference study. Evaluation images are generated with COCO30K prompts with a resolution of 2048 $\times$ 2048.}
\label{tab:ai}
% \resizebox{\linewidth}{!}{
\begin{tabular}{ccc  cc cc}
\toprule
\midrule
\bf Method/Setting & Overall Quality & Rank & Prompt Alignment & Rank & Visual Aesthetics & Rank\\
\midrule
\midrule
SDEdit & 87.09 & 2 &90.99 & 2 & 89.18 & 2\\
\rowcolor{blue!20}
w/ URAE & 88.23 & 1 &92.49 & 1 & 90.09& 1\\
\midrule
I-Max & 88.24 & 2 &91.38 & 2 & 89.96& 2\\
\rowcolor{blue!20}
w/ URAE & 89.12 & 1 & 92.58 & 1 & 90.86& 1\\
\midrule
\midrule
FLUX1.1 [Pro] Ultra & 90.42 & 1 & 93.53 & 2 & 90.42 & 2\\
\midrule
PixArt-Sigma-XL & 86.13 & 4 & 88.71 & 6 & 86.31& 5\\
Sana-1.6B & 86.46 & 3 & 90.25 & 3 & 87.80 & 3\\
FLUX-1.dev & 84.23 & 6 & 89.05 & 5 & 84.88& 6\\
FLUX-1.dev$^*$ & 86.05 & 5 & 89.48 & 4 & 87.02& 4\\
\rowcolor{blue!20}
w/ URAE & 89.71 & 2 & 93.64 & 1 & 91.47 & 1\\
\midrule
\bottomrule
\end{tabular}
% }
\end{center}
\end{table}
The designed prompt to evaluate visual aesthetics of images is as follows:
\begin{tcolorbox}[colframe=gray!70, colback=white, sharp corners]
As an AI visual assistant, you are an evaluator specialized in image quality analysis for high-resolution text-to-image generation models. When presented with a specific caption, it is required to evaluate and determine which image exhibits greater photorealism and aesthetical, in terms of clarity, richness of detail, and overall quality. Please pay attention to the key factors, including image style, the artistic quality of the image, realism, etc., the prompt is \textit{``your prompt"}. Please output [Image 1] if the first image is better, [Image 2] if the second image is better, and give me the reason.
\end{tcolorbox}
The designed prompt to evaluate the prompt alignment of images is as follows:
\begin{tcolorbox}[colframe=gray!70, colback=white, sharp corners]
As an AI visual assistant, you are an evaluator specialized in image quality analysis for high-resolution text-to-image generation models. Given a specific caption, you need to judge which image aligns with the caption more closely. Please pay attention to the key information, including object identities, properties, spatial relationships, object numbers and image style, etc., the prompt is \textit{``your prompt"}. Please output [Image 1] if the first image is better, [Image 2] if the second image is better, and give me the reason.
\end{tcolorbox}
For quantitative analysis, the designed prompt to evaluate the overall quality is as follows:
\begin{tcolorbox}[colframe=gray!70, colback=white, sharp corners]
As an AI visual assistant, you specialize in evaluating image quality for high-resolution text-to-image generation models. Given a specific caption, please evaluate the overall quality of the image by considering both content alignment and technical excellence. For content alignment, assess the key information including object identities, properties, spatial relationships, object numbers and caption-specified style. For technical quality, evaluate the image's photorealism and aesthetics, focusing on clarity, richness of detail, artistic quality, and overall visual appeal. Please analyze how well the image performs in both aspects to determine its comprehensive quality. The prompt is: \textit{``your prompt"}. Please output strictly the score from 0 to 100. Do not provide any explanation or additional text beyond this numeric score.
\end{tcolorbox}
The designed prompt to evaluate the visual aesthetics of images is as follows:
\begin{tcolorbox}[colframe=gray!70, colback=white, sharp corners]
As an AI visual assistant, you specialize in evaluating image quality for high-resolution text-to-image generation models. When given a specific caption, you are required to assess the image and assign a 0-100 score, reflecting its photorealism, aesthetic appeal, clarity, richness of detail, and overall quality. Key factors to consider include image style, artistic quality, and realism. The prompt is: \textit{``your prompt"}. Please output strictly the score from 0 to 100. Do not provide any explanation or additional text beyond this numeric score.
\end{tcolorbox}
The designed prompt to evaluate the prompt alignment of images is as follows:
\begin{tcolorbox}[colframe=gray!70, colback=white, sharp corners]
As an AI visual assistant, you are an evaluator specialized in image quality analysis for high-resolution text-to-image generation models. Given a specific caption, you need to determine the score 0-100 that the image aligns with the caption. Please pay attention to the key information, including object identities, properties, spatial relationships, object numbers and image style, etc., the prompt is \textit{``your prompt"}. Please output strictly the score from 0 to 100, reflecting how accurately the image aligns with the caption. Do not provide any explanation or additional text beyond this numeric score.
\end{tcolorbox}

\subsection{Prompts Used in Main Manuscript}

1. \textit{girl with pink hair, vaporwave style, retro aesthetic, cyberpunk, vibrant, neon colors, vintage 80s and 90s style, highly detailed.}\\
2. \textit{Astronaut in a jungle, cold color palette, muted colors, detailed, 8k.}\\
3. \textit{A whimsical village scene is nestled within an enormous teacup, where winding cobblestone streets and quaint cottages create a surreal microcosm reminiscent of dreamlike landscapes. From a high vantage point, the viewer looks down on the diminutive inhabitants going about their day-to-day activities, contrasting the juxtaposition of innocence and chaos in this fantastical setting. Soft, ethereal lighting envelops the scene, creating an atmosphere of tranquility as ordinary life intertwines with elements of fantasy and absurdity to weave a captivating visual narrative that invites the audience into an extraordinary world where the commonplace becomes surreal.} \\
4. \textit{A surreal landscape depicting an ethereal fusion of natural beauty and fantastical architecture, reminiscent of Salvador Dali's dreamlike paintings. From above the clouds, one gazes upon a colossal tower emerging from the earth, its intricate gears visible as it merges seamlessly with a tranquil mountain lake. The scene is bathed in an otherworldly glow, casting lavender and gold hues across the sky, while delicate cherry blossoms flutter gently in the foreground, adding a sense of serenity to this breathtaking vision where time and nature intertwine.}\\
5. \textit{Imagine an enchanting scene where the dreamy allure of Klimt's celestial brushstrokes intertwines with the magical wonder of Studio Ghibli animations, showcasing a mesmerizing underwater realm teeming with life. In this captivating portrait, a luminous mermaid princess emerges from a vibrant coral landscape, her iridescent silver hair flowing like cascading waterfalls as she sings a hauntingly beautiful melody that resonates through the depths. From an aerial perspective, the scene unfolds in a surreal and awe-inspiring manner, capturing both the ethereal grace of this mythical creature and the breathtaking beauty of her aquatic surroundings, creating a harmonious blend of fantasy and reality that invites viewers to lose themselves in its enchanting embrace.}\\
6. \textit{A whimsical portrait of an otherworldly fairy with luminescent wings captures her enchanting features from an unconventional angle beneath a floating moonlit garden. Rendered in high-definition digital art, her luminous eyes gleam like stars and her flowing hair seems caught in an unseen breeze, while the intricate details of shimmering scales on her wings and petals entwined within her tresses contribute to the overall sense of wonder. The soft focus of distant flowers and glistening dewdrops engulfs the scene in a dreamy atmosphere, harmoniously blending with the fairy's ethereal presence in this captivating tableau that evokes feelings of enchantment and wonder.}\\
7. \textit{Steampunk airship floating above a misty Victorian cityscape, intricate brass and copper mechanical details, golden hour lighting, billowing clouds, detailed architectural elements, rich warm color palette, cinematic composition.}\\
8. \textit{Create an image in the surrealistic style, capturing a unique bird's-eye view of a whimsical scene where a tall giraffe stands atop an oversized piano in a lush garden filled with oversized, vibrant flowers. The giraffe appears to play an invisible melody on the keys, its long neck bending gracefully over the keyboard while its feet dance rhythmically on the pedals. This striking juxtaposition should evoke a sense of wonder and enchantment, blending elements of natural beauty with unexpected musical elements in a dreamlike atmosphere.}\\
9. \textit{A sleek black luxury sedan parked on a rain-soaked city street at night, reflecting neon lights from nearby buildings. The wet pavement glistens, and the car's smooth curves are highlighted by the ambient glow of the urban environment.}\\
10. \textit{Barbarian woman riding a red dragon, holding a broadsword, in gold armour.}\\
11. \textit{A person wearing a Spider-Man suit in the game Half-Life Alyx.}\\
12. \textit{A person staring into a lucid dream world with an adventure waiting.}\\
13. \textit{A dreamlike landscape depicting an ethereal giraffe with elongated limbs and neck gracefully floating above a surreal desert oasis. This captivating scene is captured from a low-angle perspective, evoking a sense of wonder as the giraffe appears to defy gravity in a whimsical juxtaposition of elements. The image combines vibrant colors and melting textures to create an imaginative and thought-provoking vision that blurs the lines between reality and fantasy, inviting viewers to explore the depths of their own imagination.}\\
14. \textit{Craft an image in the surreal digital art style, depicting a dreamlike portrait of a young woman whose face merges with a complex floral arrangement. This scene is viewed from an elevated perspective, giving it a unique `bug's-eye view', making her appear to be enveloped within a grand garden teeming with vivid, distorted blooms emerging directly from her body and hair. The composition should inspire awe and intrigue as the petals weave into her facial contours, creating a fantastical union of nature and human form. A gentle, diffused lighting bathes this captivating scene, casting delicate shadows that dance across the textures of both the flowers and her translucent skin.}\\
15. \textit{An enigmatic dream landscape is captured in this whimsical scene, where an intricate tower with distorted, fluid architecture dominates the horizon. The sky above is filled with floating islands, their shapes shifting as if viewed through a fisheye lens, emphasizing the surreal nature of the setting. In the foreground, a curious child stands on one of these drifting lands, wide-eyed and flowing-haired, gazing up at the enigmatic tower with a mix of awe and unease, evoking an emotional spectrum that ranges from curiosity to uneasiness in this fantastical landscape.}\\
\subsection{Prompts Used in Appendix}
1. \textit{A portrait of an enigmatic cyberpunk warrior woman, her sleek black hair cascading behind her as she holds a high-tech sword in one hand and a futuristic communication device in the other. Standing atop a towering skyscraper at sunset, she is illuminated by the warm glow of city lights, her armor reflecting the vibrant hues of the sky. The scene is rendered with dramatic lighting that accentuates both the sharp angles of her ancient-style attire and the sleek curves of her modern technology, conveying an atmosphere of awe and wonder at the fusion of old and new. Her expression combines fierce determination with thoughtful contemplation, symbolizing the collision of advanced civilization and traditional wisdom she embodies. This striking digital art piece captures the essence of a warrior existing between worlds, bridging the gap between the past and future.}\\
2. \textit{An elven queen wearing transparent silk in a fantasy character portrait.}\\
3. \textit{A busy fantasy street depicting a single street within an old city lined with quirky shops, old buildings, cobblestones, and street life.}\\
4. \textit{A young engineer man with cybernetic enhancements wearing a suit and bowtie, a detailed mask, and a gloomy expression, with half of his face mechanical.}
5. \textit{Visualize an enchanting amalgamation of impressionist serenity and fantastical landscapes, where a graceful Japanese maiden in a traditional kimono gracefully traverses a serene pond filled with iridescent water lilies. The petals of these delicate flowers shimmer under the golden hour light, radiating a spectrum of soft pastel colors that envelop the scene in an atmosphere of tranquility. As she moves through this dreamlike environment, her reflection dances across the water's surface, creating a mesmerizing interplay of color and movement that captures both the timeless elegance of impressionist artistry and the enchanting magic of cinematic storytelling in one captivating masterpiece.}\\
6. \textit{((full body)) looking at sky)),Quality, looking at viewer, lot viewer, sense of depth and dimenck, making the woman's f more. The overall mood of the image is mysterious and ethereal. anime style, intricate detail.}\\
7. \textit{A captivating digital artwork in a surrealistic style, featuring an alluring ballet dancer gracefully suspended above a dreamlike landscape. The ethereal dancer's movements are captured from a low-angle perspective, highlighting her elegant limbs and flowing attire, reminiscent of a fish-eye lens. The surreal backdrop is a melting, impressionist-inspired cityscape with fluid architecture that seamlessly blends into the sky, creating an atmospheric mood of wonder and enchantment. This mesmerizing scene balances between reality and fantasy, inviting viewers to explore the dreamlike world of the ballet dancer and immerse themselves in the whimsical beauty of her performance.}\\
8. \textit{kodak film potrait , girl surrounded with bubbles detailed, dramatic lighting shadow (lofi, analog-style).}\\
9. \textit{Imagine a futuristic digital art depiction of Earth at dusk, where towering skyscrapers intertwine with flourishing vertical gardens in an intricate fusion of nature and technology. The scene is captured from a unique vantage point, reminiscent of a 'bug's-eye view', as the sky transitions from golden hues to deep purples and blues, casting elongated shadows across the urban landscape. Amidst this bustling metropolis, a solitary figure draped in ethereal, flowing robes stands atop one of the tallest structures, their silhouette softened by the waning light and embraced by the verdant surroundings. This enigmatic character holds an antiquated device that emanates a gentle, pulsating glow, contrasting with the cool metallic surfaces of the cityscape and evoking a sense of reverence for the past and wonder about the future.}\\
10. \textit{Imagine an otherworldly coastal view in a dreamlike style reminiscent of surrealism, where a grand dragonfly elegantly drifts above a peaceful aquatic expanse. The mirrored sky on its surface forms a captivating mirage as if one could submerge into the cosmos themselves. This mesmerizing scene is captured from a unique 'insect's-eye perspective', revealing intricate patterns of the dragonfly's dainty wings and vivid hues, while a gentle, amber twilight bathes the tranquil tableau in a warm embrace, stirring emotions of tranquility and wonder simultaneously.}\\
11. \textit{Envision a dreamlike landscape where surrealism intertwines with fantasy. In this whimsical realm, instead of traditional timepieces, floating islands crafted from diverse cheeses take the place of melting clocks. These cheese islands are vibrant ecosystems teeming with peculiar creatures reminiscent of those found in Alice's adventures. A giant, levitating chessboard adds to the fantastical setting as its pieces come alive and engage in playful interactions with the island inhabitants. This aerial view captures the essence of a surreal dreamscape that blends Dali-esque surrealism with Carrollian fantasy, evoking an atmosphere of wonder and curiosity about the bizarre world unfolding before our eyes.}\\
12. \textit{A high-resolution digital illustration showcases an individual standing at the edge of a colossal, intricately designed cityscape, reminiscent of steampunk architecture intertwined with ancient Mayan ruins. From a unique aerial perspective, this expansive panorama reveals towering mechanical spires and cogs, juxtaposed against a vivid azure sky. As the sun sets, its golden rays cast elongated shadows over the labyrinthine structures, instilling a sense of wonder and intrigue. The protagonist, an enigmatic figure wearing a sleek, modern exoskeleton, stands poised on the brink of this massive metropolis, one foot stepping forward as if ready to unveil the secrets hidden within its gears and mechanisms. This mesmerizing image skillfully fuses elements of science fiction, ancient mythology, and futuristic technology in a visually captivating composition that transports viewers into an atmospheric world brimming with both awe and mystery.}

\subsection{Prompts Used in User Study}\label{user}
1. \textit{Craft an image depicting a surreal dreamscape with a majestic unicorn floating amidst tumultuous waves, viewed from an aerial perspective akin to observing through a bird's eyes soaring above the sea. This scene captures both serene beauty and chaotic turbulence in one fantastical landscape. Utilize vibrant, contrasting colors, featuring deep blues for the stormy sea and fiery oranges and purples for the swirling clouds overhead, creating an emotional gradient that evokes wonder, danger, and ethereal grace simultaneously.}\\
2. \textit{A captivating Art Nouveau-inspired image showcases a celestial enchantress gracefully dancing amidst a swirling vortex of shimmering stardust, her ethereal gown intricately woven with delicate silver threads reminiscent of cosmic nebulae. Captured from an elevated perspective that accentuates the vastness and grandeur of the cosmos, this scene radiates a sense of wonder, enchantment, and serenity as it invites viewers to marvel at the luminous beauty of the muse against the backdrop of the infinite expanse.}\\
3. \textit{A surreal digital artwork depicting an enigmatic floating cityscape composed of inverted ziggurats suspended in midair. From an aerial perspective, the city appears to hover over an endless expanse of rippling water. As one draws closer to the water's surface, the reflection reveals a mirrored image of the city, but with its architecture twisted and distorted by the shifting tides. The vibrant colors and intricate details evoke a sense of wonder mixed with unease, inviting the viewer to contemplate the relationship between reality and illusion.}\\
4. \textit{A surreal digital artwork depicting a bustling futuristic cityscape at night, with towering skyscrapers adorned in vibrant, abstract shapes. The scene transitions from sharp clarity to a soft, dreamlike atmosphere as it approaches the horizon, evoking both awe and uncertainty in the viewer. Neon lights pulse within the city, seemingly melting and warping in the air, creating mesmerizing patterns and reflections on the wet streets below. From a high-altitude perspective, this enigmatic metropolis is captured in an aerial view, inviting contemplation of the convergence of reality and dreams.}\\
5. \textit{In the style of visionary art, depict a serene female figure draped in a luminous white gown with intricate mandala patterns in deep blue and vibrant teal hues. This ethereal portrait is viewed from an aerial perspective, showcasing the subject against a cosmic background that seamlessly blends into swirling galaxies and nebulae. The radiant colors and harmonious compositions evoke a profound sense of spiritual awakening and interconnectedness with the vast universe around us.}\\
6. \textit{In the style of visionary art, depict a serene female figure draped in a luminous white gown with intricate mandala patterns in deep blue and vibrant teal hues. This ethereal portrait is viewed from an aerial perspective, showcasing the subject against a cosmic background that seamlessly blends into swirling galaxies and nebulae. The radiant colors and harmonious compositions evoke a profound sense of spiritual awakening and interconnectedness with the vast universe around us.}\\
7. \textit{A dreamlike landscape emerges from a first-person viewpoint, immersing the observer in an alluring world where waterlilies of soft lavender and violet hues gracefully drift on the surface of an opalescent pond. Towering lotus blossoms stretch towards an indigo sky embellished with celestial bodies that gleam like stars, invoking both tranquility and awe. A regal swan presides over this fantastical garden, its iridescent feathers creating captivating ripples across the water that seem to distort time itself, crafting a harmonious melody of dreams and nature, encapsulating the spirit of beauty and whimsy in one stunning tableau.}\\
8. \textit{Envision an otherworldly aquatic environment where a graceful mermaid adorned with iridescent scales reminiscent of deep-sea hues gracefully dances amidst vibrant coral formations. Her tranquil expression mirrors a state of contemplative introspection, as if she is enveloped in the enigmatic depths from an unconventional vantage point – that of a diminutive seashell. Delicate intricacies emerge, such as her cascading tresses mirroring tender seaweed and how sunlight weaves through the water to cast enchanting patterns upon her skin. This captivating scene instills a sense of awe and reflection while preserving an ethereal aura reminiscent of surrealistic art.}\\
9. \textit{Craft an enchanting surrealist scene showcasing a 'bug's-eye view' perspective of a chess game occurring on a shifting landscape. In this scene, a majestic phoenix perches atop a black bishop, its vibrant wings casting intricate shadows across the checkerboard expanse below. The background alternates between lush tropical forests and barren deserts with each move made by the ethereal beings participating in this mysterious match, instilling feelings of intrigue and wonder as they traverse the unpredictable terrain under the eerie illumination of a full moon.}\\
10. \textit{Imagine an enchanting digital artwork depicting a dragonfly's perspective above a tranquil pond, reminiscent of Monet's captivating water lilies. The scene showcases lush, vibrant vegetation surrounding the serene water surface, with an emotional gradient highlighting the beauty and enchantment as light gracefully dances across the composition. Merging elements of impressionism with high-resolution photorealistic textures, this piece evokes a sense of awe and wonder, creating an ethereal atmosphere that captures the dragonfly's mesmerizing flight amidst a blooming floral paradise.}\\
11. \textit{Imagine a lively digital painting capturing the exhilarating spirit of an anime-inspired hoverboard race in a neon-drenched cyberpunk cityscape at twilight. The scene pulses with action as diverse characters navigate through a maze-like urban landscape of towering skyscrapers and radiant billboards, deftly maneuvering around airborne vehicles and zipping pedestrians while leaving trails of shimmering pixels behind them. As the sun dips below the horizon, its dramatic lighting casts elongated shadows that accentuate the futuristic architecture and high-tech trinkets sprinkled throughout, creating a rich tapestry of cool blues, purples, and pinks that intensify the emotional stakes and anticipation as the race nears its peak. This captivating image, with its dynamic composition and vibrant use of color and light, transports viewers into an enthralling world where technology and nature intertwine in a dazzling display of innovation and style.}\\
12. \textit{An enigmatic digital artwork showcases an celestial ballerina gracefully spinning across the cosmos, her luminescent dress shimmering with iridescence against the inky black backdrop of space. The scene is observed from a unique 'worm's eye view', accentuating the grandeur and elegance of this otherworldly dancer as she whirls amidst nebulae and stars, creating an entrancing spectacle that captivates both the senses and the imagination while evoking a sense of wonder and awe for the hidden beauty within the universe.}\\
13. \textit{A mesmerizing digital artwork portraying an enchanting celestial sorceress in shimmering robes adorned with starlight-inspired patterns. Seen from above, her captivating gaze reflects the light of distant galaxies as she skillfully shapes the cosmos with her mystical staff, creating swirling nebulae and brilliant constellations in a breathtaking display. The backdrop showcases a deep purple expanse of space filled with nebulous clouds, pulsating stars, and enigmatic planets, generating an emotional gradient that balances wonder and mystery within this celestial tableau.}\\
14. \textit{Craft an image that embodies spiritual realism with celestial elements, depicting an astronaut in a reflective spacesuit adorned with intricate mandala patterns. This ethereal figure floats amidst the cosmic void, traversing through a luminous wormhole that connects our physical world to a higher plane of existence. The astronaut's journey is captured from an 'eye-of-the-needle' perspective, enveloped by vivid colors and profound symbolism that evoke wonder, transcendence, and spiritual awakening in the vast expanse of space.}\\
15. \textit{A celestial ballet unfolds within an ethereal nebula as galaxies collide in a mesmerizing dance of cosmic forces. From an impossible vantage point, suspended above the event horizon, we gaze upon this breathtaking spectacle. The scene captures the chaotic beauty and sublime mystery of the universe, with swirling patterns of stars, nebulas, and cosmic dust illuminated by an otherworldly light source that casts long shadows across the celestial plane. This visually stunning composition evokes a sense of awe and wonder at the sheer scale and complexity of our cosmos, as well as the delicate balance between order and chaos in the grand design. Inspired by the visionary artistry of Escherian architecture and the cosmic explorations of space telescopes, this image invites viewers to ponder the infinite mysteries that lie beyond the stars.}\\
16. \textit{Create a dreamlike still life scene blending impressionistic water lilies with surreal melting elements. Picture a tranquil garden pond adorned with softly ruffled blue, pink, and green flowers, as if captured in an ethereal dance of light and color. In the background, a tower melts into the distant horizon, its hands frozen in time, evoking a sense of timeless wonder. The reflection on the water's surface distorts the landscape into mesmerizing shapes and colors, blurring the line between reality and fantasy, enveloping the viewer in an enchanting atmosphere of surreal beauty and awe.}\\
17. \textit{An awe-inspiring image reveals an enigmatic blend of swirling celestial patterns reminiscent of van Gogh's iconic brushstrokes and the surreal melting clocks synonymous with Dali's dreamscapes. Set within the opulent interior of a vast library, the scene captivates the viewer's gaze as they stand amidst towering shelves adorned with golden-hued spines. Above, an intricate celestial map unfolds across the ceiling, its constellations echoing van Gogh's dynamic style while clocks dissolve into nebulous forms that cast whimsical shadows over the polished marble floors. This surreal tableau masterfully combines wonder and introspection, inviting the observer to embark on a journey through time and space within its harmonious visual symphony.}\\
18. \textit{A captivating digital artwork merges Victorian-era street market with an underwater world, creating a surreal fusion of reality and fantasy. The bustling activity of vendors selling exotic goods under a sky of swirling auroras transitions seamlessly into the vibrant depths of the ocean, inviting viewers to explore this mysterious realm. This unique blend of surface-level excitement and serene tranquility evokes curiosity and wonder, guiding the audience through a mesmerizing journey that defies traditional boundaries.}\\
19. \textit{An enchanting digital artwork portrays an underwater haven where fantastical creatures interact harmoniously in a style reminiscent of James Gurney's Dinotopia. From an aerial perspective, this captivating scene reveals the gentle touch between a regal triceratops and a graceful mermaid, as they share a tender moment under the luminous sunlight filtering through the pristine waters. This mesmerizing encounter evokes wonder and tranquility, skillfully blending fantasy with prehistoric elements to create a visually stunning tableau that transcends time and imagination.}\\
20. \textit{Craft an enchanting digital artwork that fuses surrealism with vibrant colors, depicting an underwater realm where fish gracefully perform ballet in harmony with the flowing currents. This unique perspective showcases a dreamlike dance between elegant sea creatures and floating bubbles, bathed in soft pastel hues evoking a sense of wonder and tranquility. Delicate brushstrokes and fantastical shapes intertwine to create a captivating visual symphony, inviting viewers into an otherworldly realm where reality and imagination seamlessly blend.}

\section{More Experimental Results}
\subsection{More Results on AI Preference Study}\label{ai-study}
We additionally use GPT-4o for a quantitative analysis, allowing for a more intuitive evaluation of performance of each method across different assessment dimensions. Our experimental setting and pairwise comparison remain consistent. We randomly select 300 prompts from the COCO30K dataset and generate images of 2048$\times$2048. The results are shown in Table~\ref{tab:ai}. The results indicate that our method achieves an exceptionally high level across all three dimensions and is comparable to SOTA model FLUX1.1 [Pro] Ultra.

\subsection{More Qualitative Examples}\label{more-qualitative}
\begin{figure}[t]
  \centering
   \includegraphics[width=0.85\linewidth]{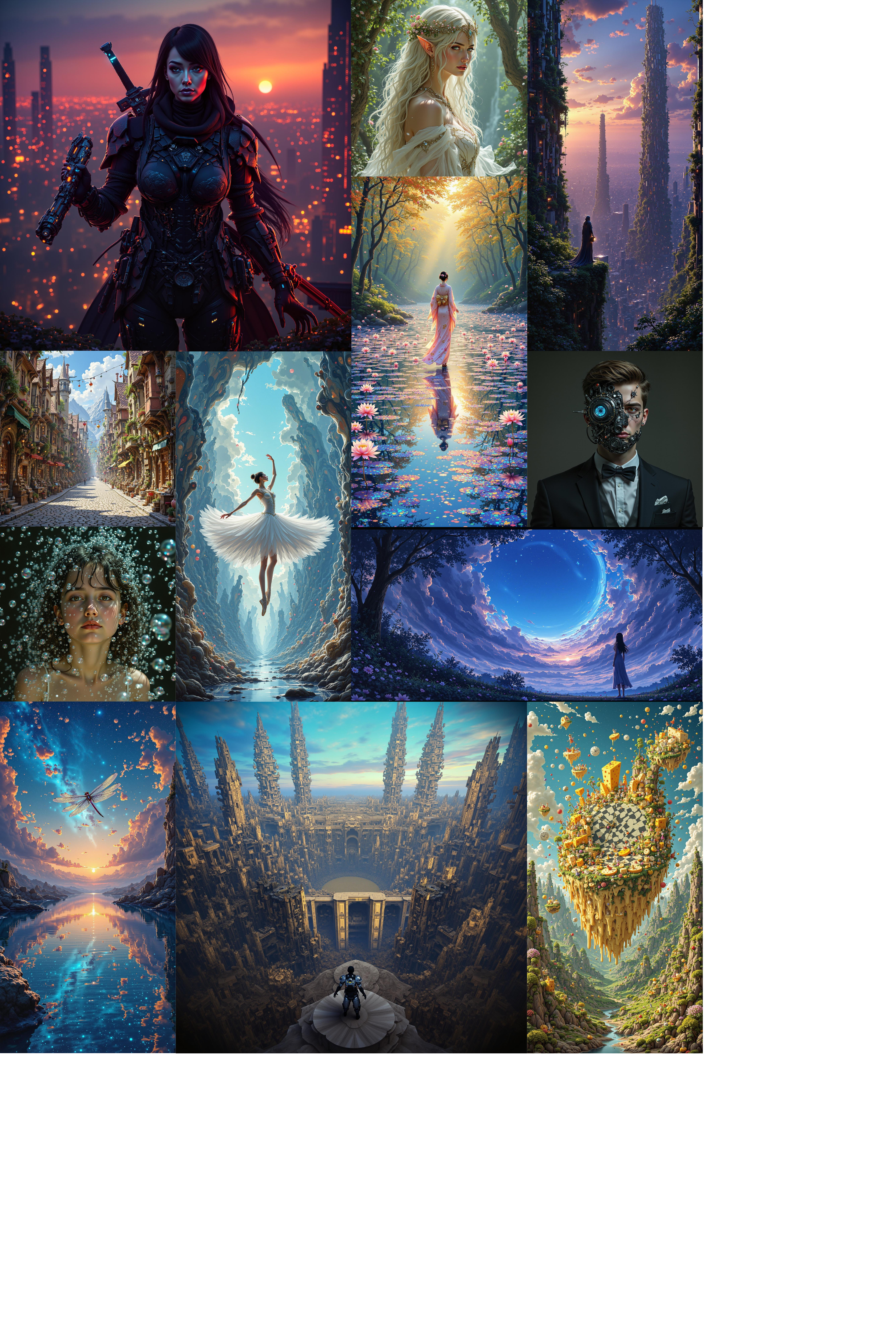}
   \caption{More high-resolution results of our proposed method. The image in upper left is of 4096 $\times$ 4096.}  
   \label{fig:gpt-app}
\end{figure}

We include more qualitative samples in Fig.~\ref{fig:gpt-app} to further demonstrate the superior performance on handling detailed and complex structures and textures in ultra-resolution image generation. 

%%%%%%%%%%%%%%%%%%%%%%%%%%%%%%%%%%%%%%%%%%%%%%%%%%%%%%%%%%%%%%%%%%%%%%%%%%%%%%%
%%%%%%%%%%%%%%%%%%%%%%%%%%%%%%%%%%%%%%%%%%%%%%%%%%%%%%%%%%%%%%%%%%%%%%%%%%%%%%%

\end{document}